\documentclass{article}

\usepackage{amsfonts}
\usepackage{amssymb}
\usepackage{amsmath}
\usepackage{amsthm}
\usepackage{latexsym}
\usepackage{verbatim}
\usepackage{epsfig}
\usepackage{amsbsy}
\usepackage{amscd}
\usepackage{bm}

\usepackage{graphicx}

\usepackage{ifthen} 
\usepackage{xcolor}
\usepackage[colorlinks=true, linkcolor=teal, anchorcolor=teal,
            citecolor=blue,  filecolor=blue, menucolor=blue, pagecolor=teal,											urlcolor=blue, plainpages=false, pdfpagelabels]{hyperref}

\newcommand{\mymail}[1]{\href{mailto:#1}{\texttt{#1}}}

\usepackage{fancyhdr}
\usepackage{currfile}
\fancypagestyle{firststyle}{

\cfoot{{\footnotesize \texttt{\currfilename}, version: \texttt{\today} }} 
}

\newcommand{\setauthA}[1]{\def\authA{#1}}
\newcommand{\setauthB}[1]{\def\authB{#1}}
\newcommand{\setauthC}[1]{\def\authC{#1}}

\def\printA{\begin{tabular}{l} \authA \end{tabular}}
\def\printB{\begin{tabular}{l} \authB \end{tabular}}
\def\printC{\begin{tabular}{l} \authC \end{tabular}}

\newcommand{\makemytitle}[1]{\begin{center}{\textsf{\LARGE #1}}
  \end{center}
}

\usepackage[sf,bf]{titlesec}

\usepackage{algorithmic}
\usepackage{algorithm}

\usepackage[nonamebreak,square,numbers,sort]{natbib}

\usepackage[letterpaper, textheight = 676pt]{geometry}
\providecommand{\M}[1]{\mathbf#1}
\providecommand{\mc}[1]{\mathcal#1}
\providecommand{\mc}[1]{\mathcal#1}

\newcommand{\R}{{\mathbb R}}

\DeclareMathOperator{\E}{\mathbf{E}}
\DeclareMathOperator{\p}{\mathbf{P}}

\DeclareMathOperator{\tr}{tr}

\providecommand{\T}{\top} 

\DeclareMathOperator*{\argmin}{argmin}
\DeclareMathOperator*{\argmax}{argmax}

\providecommand{\wt}[1]{\widetilde{#1}}
\providecommand{\wh}[1]{\widehat{#1}}

\providecommand{\nnorm}[1]{ \lVert#1 \rVert}
\newcommand{\scp}[2]{\left\langle#1, #2\right\rangle}
\newcommand{\nscp}[2]{\langle#1, #2\rangle}

\newcommand{\blanco}[1]{  }

\newcommand{\deriv}[3]{%
\ifthenelse{#1 = 1}{\frac{d\,#2}{d\,#3}}{\frac{d^{{#1}} #2}{d{#3}^{{#1}}}}
}

\newcommand{\partials}[3]{%
\ifthenelse{#1 = 1}{\frac{\partial\,#2}{\partial\,#3}}{\frac{\partial^{#1}
    #2}{\partial#3^{#1}}}
} 

\def\su{\sum_{i=1}^n}

\def \coloneq{\mathrel{\mathop:}=}
\def \invcoloneq{=\mathrel{\mathop:}}
\def \eps{\varepsilon}

\newtheorem{theo}{Theorem}
\newtheorem{propo}{Theorem}

\newtheorem{prop}[propo]{Proposition}

\usepackage{url}
\usepackage{tikz}
\def\R{\mathbb{R}}
\def\tr{\mathrm{tr}}

\def\x{\mathbf{x}}
\def\y{\mathbf{y}}
\def\eps{\epsilon}

\def\dH{d_{\textsf{H}}}

\usepackage{enumitem}
\usepackage{subfigure}
\newboolean{isjournal}
\setboolean{isjournal}{true}

\newboolean{nocolors}
\setboolean{nocolors}{false}

\newboolean{usemtpro}
\setboolean{usemtpro}{false}

\ifthenelse{\boolean{nocolors}}{\hypersetup{colorlinks=false}}{}

\makeatletter
\newcommand\footnoteref[1]{\protected@xdef\@thefnmark{\ref{#1}}\@footnotemark}
\makeatother

\setauthC{{\bfseries Martin Slawski}$^*$}

\setauthB{{\bfseries Emanuel Ben-David$^{\dagger}$}}

\setauthA{{\bfseries Zhenbang Wang}$^*$}

\begin{document}
\thispagestyle{firststyle}

\makemytitle{{\Large {\bfseries {Regularization for Shuffled Data Problems via Exponential Family Priors on the Permutation Group}}}}
\vskip 3.5ex
%
{\large\begin{center}
\printA
\printB
\printC
\vskip1.5ex
{\scriptsize $^{*}$Department of Statistics, George Mason University, Fairfax, VA 22030, USA $\; \; \;$}\\[.5ex] 
\scriptsize{$^{\dagger}$Center for Statistical Research and Methodology (CSRM), U.S. Census Bureau, Suitland, MD 20746, USA}  \\[.5ex]
\mymail{zwang39@gmu.edu} \hspace*{2ex} \mymail{emanuel.ben.david@census.gov} \hspace*{2ex} \mymail{mslawsk3@gmu.edu}. 
\end{center}}
\vskip 3.5ex

\begin{abstract}
In the analysis of data sets consisting of $(X, Y)$-pairs, a tacit assumption is that each pair corresponds to the same observation unit. If, however, such pairs are obtained via record linkage of two files, this assumption can be violated as a result of mismatch error rooting, for example, in the lack of reliable identifiers in the two files. Recently, there has been a surge of interest in this setting under the term ``Shuffled data" in which the underlying correct pairing of $(X, Y)$-pairs is represented via an unknown index permutation. Explicit modeling of the permutation tends to be associated with substantial overfitting, prompting the need for suitable methods of regularization. In this paper, we propose a flexible exponential family prior on the permutation group for this purpose that can be used to integrate various structures such as sparse and locally constrained shuffling. This prior turns out to be conjugate for canonical shuffled data problems in which the likelihood conditional on a fixed permutation can be expressed as product over the corresponding $(X,Y)$-pairs. Inference is based on the EM algorithm in which the intractable E-step is approximated by the Fisher-Yates algorithm. The M-step is shown to admit a significant reduction  from $n^2$ to $n$ terms if the likelihood of $(X,Y)$-pairs has exponential family form as in the case of generalized linear models. Comparisons on synthetic and real data show that the proposed approach compares favorably to competing methods.
\end{abstract}

\section{Introduction}\label{sec:intro}
Shuffled data problems refer broadly to situations in which the goal is to perform inference for a functional
of the joint distribution of a pair of random variables $(\x,\y)$ (such as, e.g., their covariance) based on 
separate samples $\{ \x_i \}_{i = 1}^n$ and $\{ \y_i \}_{i = 1}^m$ that involve matching pairs 
$\{ (X_{\pi^*(i)}, Y_i) \}_{i = 1}^{m}$ pertaining to the same statistical unit, where
the map $\pi^*: \{1,\ldots,m\} \rightarrow \{1,\ldots,n \}$ may only be observed incompletely. This
is a rather common scenario in data integration problems in which different pieces of information 
about a shared set of entities reside in multiple data sources that need to be combined in order
to perform a given data analysis task. The process of identifying matching parts across
two or more files is often far from trivial in the absence of unique identifiers, and has
thus grown into a vast and active field of research known as \emph{record linkage} (e.g., 
\cite{Binette2020}). The above shuffled data model in terms of the unknown map $\pi^*$ represents a direct approach 
to account for mismatches in record linkage and their impact on downstream data analysis. 
Historically, shuffled data problems were first systematically discussed in a series of 
papers by DeGroot \emph{et al}.~\cite{DeGroot1971, DeGroot1976, DeGroot1980, Goel1975}, with 
little to no progress until only a few years ago given advances in computation \cite{Gutman13}. Recently,
shuffled data problems have generated much more widespread interest, fueled by novel applications 
in engineering and computer vision, among others \cite{Unnikrishnan2015, Pananjady2016, Pananjady2017}. 
Several papers have investigated the statistical limits of signal estimation and permutation
recovery in \emph{unlabeled sensing} in which the goal is to recover a signal $\theta^*$ from $n$ noisy linear
measurements $y_i = \nscp{\x_{\pi^*(i)}}{\theta^*} + \eps_i$, $1 \leq i \leq n$, where 
$\pi^*$ is an unknown index permutation \cite{Unnikrishnan2015, Pananjady2016, Hsu2017, Abid2017, TsakirisICML19}. Another 
line of research has studied similar permuted data settings in which $x$ and $y$ are scalar and related by a monotone transformation
\cite{Carpentier2016, Weed2018, Balabdoui2020, Flammarion16, Ma2020}. 

A common conclusion from these works is that
shuffled data problems are generally plagued by both statistical and computational challenges. First, 
the combinatorial nature of $\pi^*$ makes it hard to devise computationally tractable approaches with
provable guarantees. Existing algorithmic ``solutions" involve integer programming \cite{TsakirisICML19, Peng2020, Mazumder2021} and the EM algorithm \cite{Tsakiris2018,Abid2018, Gutman13}. Regardless of 
the computational challenges, shuffled data problems tend to be highly susceptible to noise and 
prone to overfitting. In fact, statistical guarantees typically involve unrealistically stringent
signal-to-noise requirements \cite{Pananjady2016, Hsu2017, SlawskiBenDavid2017}. Loosely speaking, this
issue results from the fact that the set of index permutations grows rapidly in size with $n$. This
observation suggests that suitable forms of regularization hinging on prior information on 
$\pi^*$ are needed to constrain the size of the parameter space under consideration. Several papers
consider partial shufflings in which varying fractions of $(\x_i,\y_i)$-pairs are already observed with the 
correct correspondence \cite{SlawskiBenDavid2017, Peng2021, SlawskiDiaoBenDavid2019, SlawskiBenDavidLi2019, SlawskiRahmaniLi2018, ZhangLi2020},
and only the remaining portion of the data is subject to shuffling. Another constraint commonly encountered 
in record linkage applications is that $\pi^*$ is block-structured with known composition of the blocks based on auxiliary variables that are required to agree for matching records \cite{Chambers2009, chambers2019improved, Zhang2020}. In other
applications such as signal processing and computer vision, $\pi^*$ is often constrained to act locally in the sense that 
indices are shuffled only within small time windows or image regions \cite{Ma2021, Abbasi2021}. 

The goal  of the present paper is the development of a regularization framework for shuffled data problems 
that integrates the aforementioned as well as potential other constraints in a unified fashion. For this purpose, we introduce 
an exponential family prior on the permutation group that is flexible enough to accommodate any kind
of prior information that can be expressed in terms of index pairs $(i,j)$. Conveniently, this prior
turns out to be conjugate for canonical shuffled data problems in which the likelihood conditional on a fixed permutation can be expressed as the product over the corresponding $(\x,\y)$-pairs. Inference is performed via
the Monte-Carlo EM algorithm considered in earlier works \cite{Wu1998, Abid2018, Gutman13}.  We show
that for exponential family likelihoods, the resulting M-step is particularly scalable since it only
involves $n$ instead of $n^2$ terms. Moreover,
computation of the MAP estimator of $\pi^*$ with the remaining parameters fixed is shown to reduce
to a linear assignment problem (LAP), and hence remains computationally tractable. Several theoretical results as well as a collection of experiments
for various shuffled data settings demonstrate the usefulness of regularization based on the proposed
prior in comparison to the unregularized counterpart and several other baselines. 
\vskip1.5ex 
\noindent {\bfseries Paper organization}. Section $\S$\ref{sec:approach} starts with
a detailed motivation of the proposed approach, followed by a discussion of central
technical and computational aspects. Section $\S$\ref{sec:theory} contains several
theoretical results and accompanying discussions. Numerical results on synthetic
and real data are presented in $\S$\ref{sec:experiments}. Proofs and complementary
technical details are relegated to the appendix. 

\vskip1.5ex
\noindent {\bfseries Notations}. For the convenience of the reader, an overview of the most frequently used notation throughout this paper is provided below.
{\footnotesize
\begin{center}
\begin{tabular}{|ll|ll|}
\hline
$p(\cdot)$ & density of a list of variables & $\propto$ & equality up to a positive constant \\
$p(\cdot \, | \, \cdot)$ & conditional density &  $I_n$&  $n$-by-$n$ identity matrix\\
$u \sim p$ & random variable $u$ has density $p$ & $\mathbb{I}$ & indicator function \\ 
$(\M{x}, \M{y})$ &  generic pair & $[n]$ &  $\{1,\ldots,n \}$ \\
$\mc{D} = \{ (\M{x}_i, \M{y}_i)\}_{i = 1}^n$ & observed linked data & $\pi$  & permutation on $[n]$ \\ 
$\M{X}$ & row-wise concatenation of $\{ \M{x}_i \}_{i = 1}^n$ &  $\Pi = (\pi_{ij})$                           &    corresponding permutation matrix       \\
$\M{Y}$ & row-wise concatenation of $\{ \M{y}_i \}_{i = 1}^n$ &   $\mc{P}(n)$                            &     set of permutations on $[n]$      \\
$\E_{\ldots}[]$& expectation w.r.t.~$\ldots$ & $\theta$  &  model parameter\\
$\tr$ & matrix trace & $d_{\textsf{H}}$ & Hamming distance on $\mc{P}(n)$ \\
$\nscp{A}{B}$ & $\tr(A^{\T} B)$ & $|A|$& cardinality of set $A$\\
$\nnorm{\cdot}_{\text{F}}$ & Frobenius norm & \textsf{SNR} & signal-to-noise ratio \\
\hline
\end{tabular}
\end{center}}
\noindent {\bfseries Conventions}. We often refer to a permutation via the underlying map $\pi$ and the corresponding matrix $\Pi$ in an interchangeable fashion, and accordingly 
$\mc{P}(n)$ and subsets thereof may refer to both maps and matrices. Asterisked symbols such as $\pi^*$, $\theta^*$, $\sigma_*$ etc.~refer to ground truth parameters, whereas non-asterisked symbols such as $\pi$, $\theta$, $\sigma$ etc.~refer to generic elements of the associated parameter spaces. 
\section{Approach}\label{sec:approach}
The following three subsections are dedicated to a detailed account of the approach. We start with a brief motivation before
a more formal systematic introduction and subsequent technical details pertaining to computation and model fitting. 
\subsection{Motivating examples}\label{subsec:mot_example} 
\begin{figure}
   \includegraphics[height = .18\textheight]{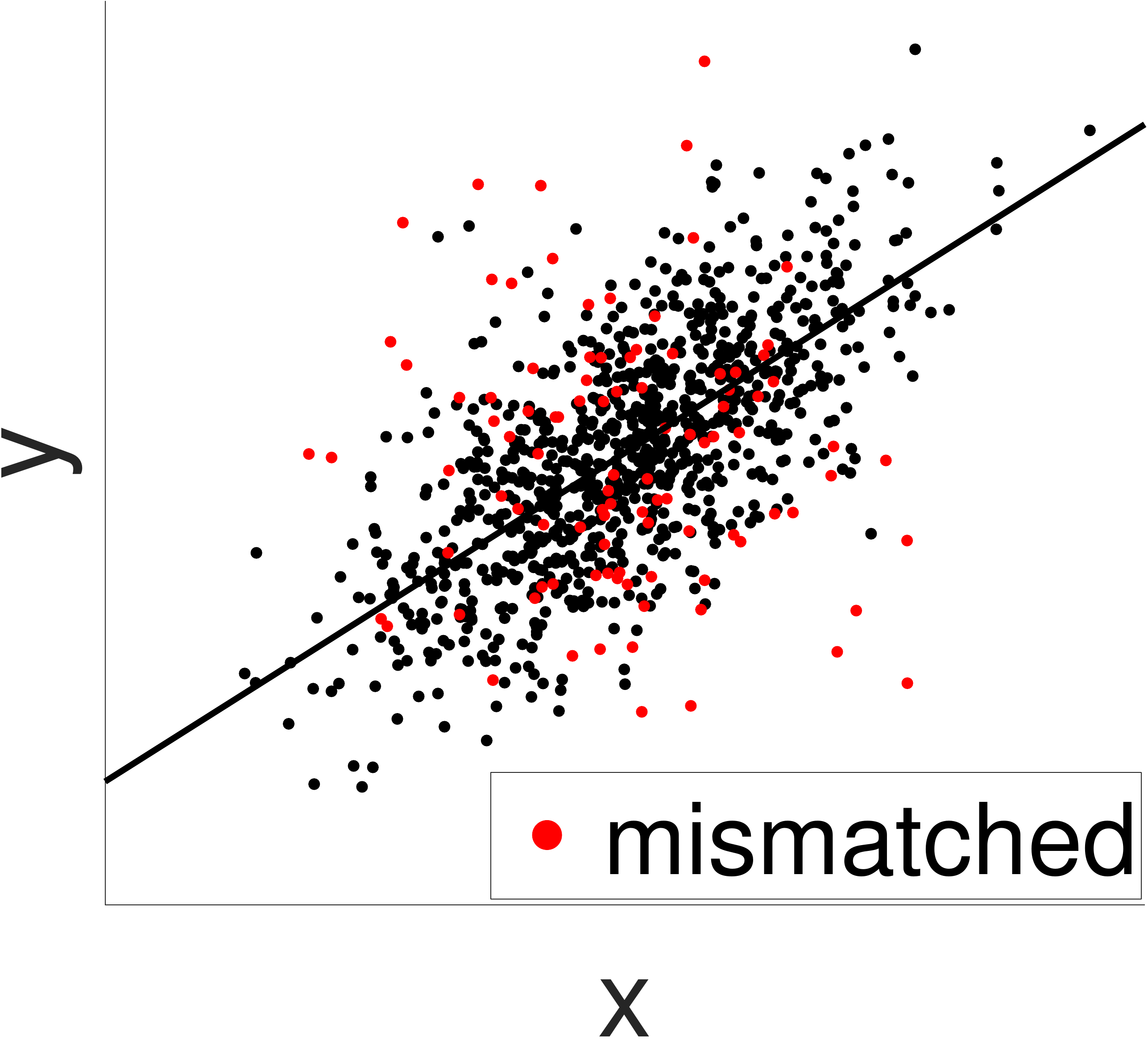}  \hspace*{1.5ex} \includegraphics[height = .18\textheight]{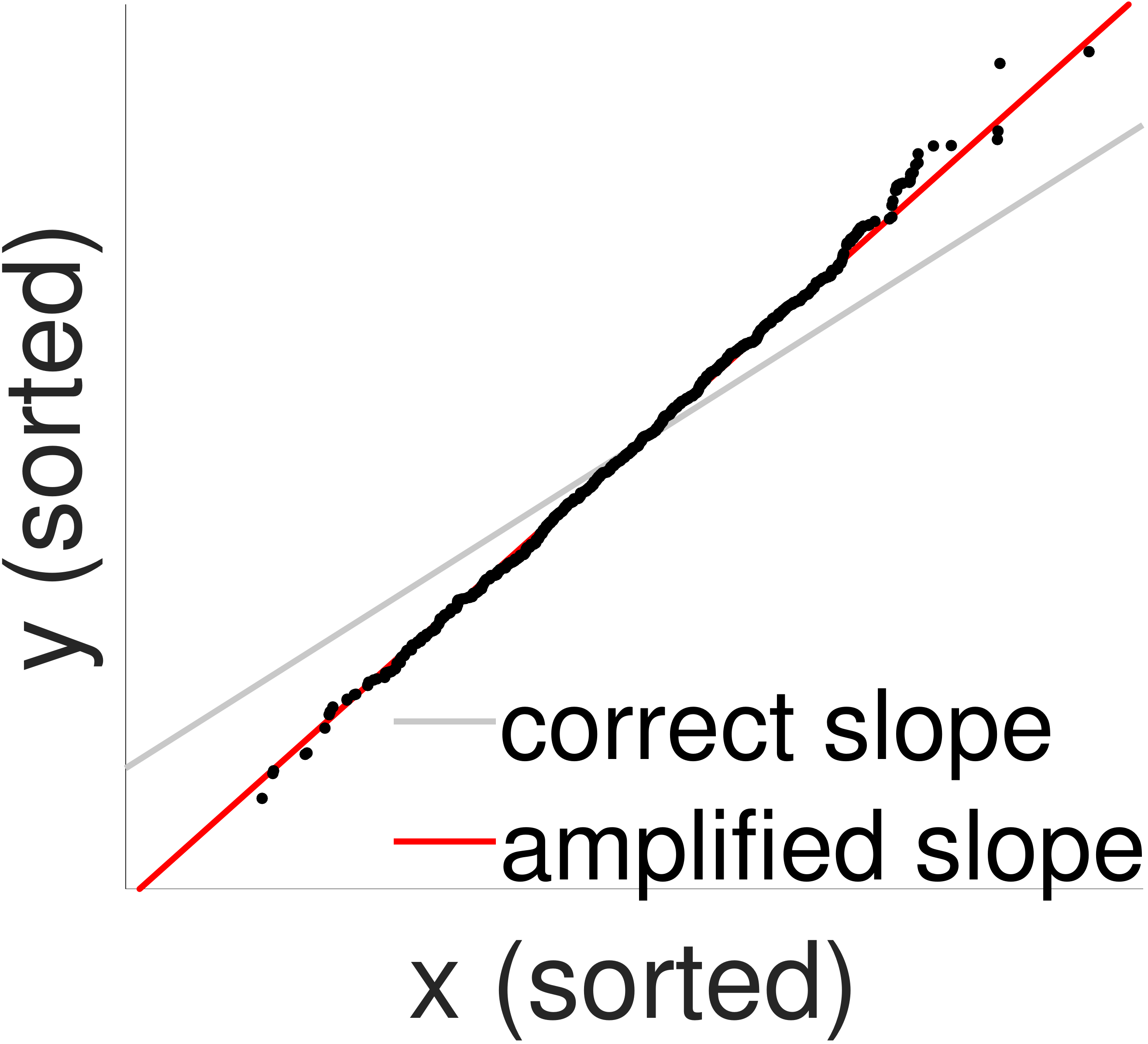} \hspace*{1.5ex} \includegraphics[height = .18\textheight]{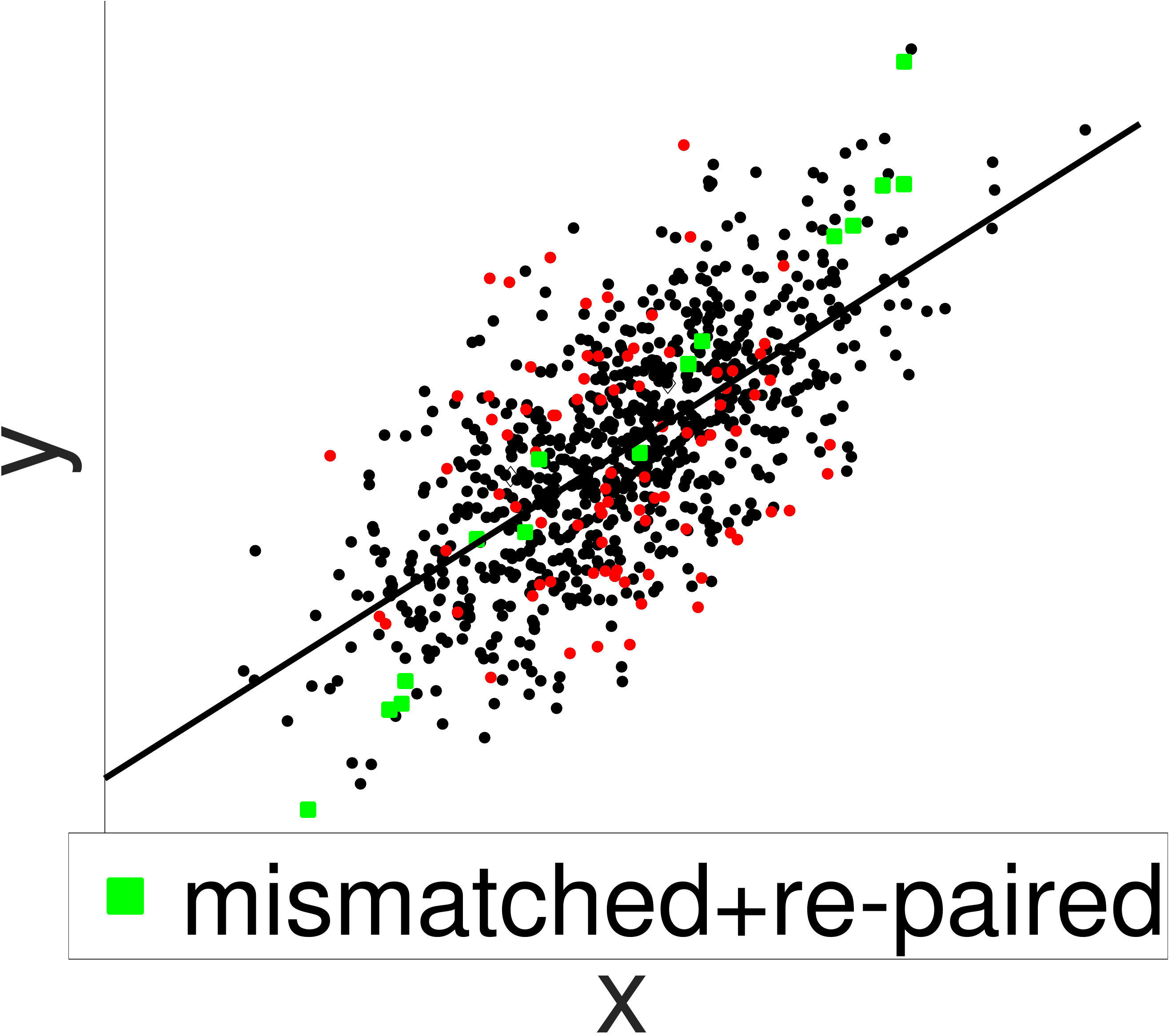}
   \vspace*{-2ex}
    \caption{L: Samples from the model $y_i = x_{\pi^*(i)} \beta^* + \eps_i$, $1 \leq i \leq n = 1,000$ as described in the text, with 10\% random mismatch. M: Re-paired data $(x_{\wh{\pi}_{\text{ML}}(i)}, y_i)_{i = 1}^n = (x_{(i)}, y_{(i)})_{i = 1}^n$ and corresponding amplified slope $\wh{\beta}_{\text{ML}}$. R: Re-paired data $(x_{\wh{\pi}(i)}, y_i)_{i = 1}^n$ based on the Hamming prior for $\pi^*$.} \label{fig:motivation_sparse}
\end{figure}

Consider the simple linear regression setup $y_i = x_{\pi^*(i)} \beta^* + \sigma_* \eps_i$, where $x_i$ and $\eps_i$ are independent standard normal random variables, $1 \leq i \leq n$, and $\pi^*$ permutes 10\% of the indices uniformly at random. Suppose that the sign of $\beta^*$ is known to be positive. Then the ML estimator of $\pi^*$ (or equivalently, the MAP estimator under a uniform prior over $\mc{P}(n)$) is given by the permutation $\wh{\pi}_{\text{ML}}$ that matches the corresponding order statistics in $\{ x_i \}_{i = 1}^n$ and $\{ y_i \}_{i = 1}^n$, i.e., 
\begin{equation}\label{eq:pihat_ML}
\su x_{\wh{\pi}_{\text{ML}}(i)} y_i  = \su x_{(i)} y_{(i)}     
\end{equation}    
As shown in Figure \ref{fig:motivation_sparse}, the estimator $\wh{\pi}_{\text{ML}}$ performs rather poorly. The scatterplot of the matching of corresponding order statistics is far from that of the underlying correct pairing. In fact, 
$\wh{\pi}_{\text{ML}}$ is associated with massive overfitting as made explicit in the sequel. Let 
\begin{equation*}
\wh{\beta}_{\text{ML}} = \su x_{(i)} y_{(i)} \Big / \sum_{i = 1}^n x_i^2, \qquad \wh{\sigma}_{\text{ML}}^2 = \frac{1}{n} \su (y_i - x_i \wh{\beta}_{\text{ML}})^2
\end{equation*}
denote the resulting ML estimators of $\beta^*$ and $\sigma_*^2$, respectively. It is straightforward to show that 
\begin{equation}\label{eq:convergence_overfitting}
\wh{\beta}_{\text{ML}} \rightarrow \sqrt{(\beta^*)^2 + \sigma_*^2}, \qquad \wh{\sigma}_{\text{ML}}^2 \rightarrow 0, \qquad \frac{1}{n} \su (x_i \beta^* - x_i \wh{\beta}_{\text{ML}})^2 \rightarrow \sigma_*^2
\end{equation}
in probability as $n \rightarrow \infty$ (see Appendix \ref{app:overfitting_ml} for a derivation). In particular, the second and third relation in \eqref{eq:convergence_overfitting} are alarming since they imply that the least squares fit absorbs all the noise.  

Figure \ref{fig:motivation_sparse} shows that the ML estimator is too aggressive in forming ``corrected" pairs
$(x_{\wh{\pi}_{\text{ML}}(i)}, y_i)$ given that only 10\% of the observations are actually mismatched, and among those 10\%, only a similarly small fraction contributes substantial mismatch that visibly exceeds the noise inherent in the problem. Sparsity of $\pi^*$ is often a reasonable assumption in post-linkage data analysis (e.g., \cite{Chambers2009, chambers2019improved, SlawskiBenDavid2017, SlawskiDiaoBenDavid2019}), where sparsity here refers to the set of mismatches $\{1 \leq i \leq n: \pi^*(i) \neq i \}$ having significantly smaller cardinality than $n$. If an upper bound on the number of mismatches, say $k$, is known, it is appropriate to consider the following constrained ML estimator of $\pi^*$: 
\begin{equation}\label{eq:sparsity_constrained}
\max_{\pi \in \mc{P}(n)} \su x_{\pi(i)} y_i \quad \text{subject to} \; \dH(\pi, \textsf{id}) \leq k,     
\end{equation}    
where $\textsf{id}$ is the identity map on $[n]$ and $\dH(\pi, \pi') = \su \mathbb{I}(\pi(i) \neq \pi'(i))$ denotes the Hamming distance between two elements $\pi, \pi'$ of $\mc{P}(n)$. For $k = n$, the maximizer of the above problem is given by $\wh{\pi}_{\text{ML}}$ and for $k < 2$, the maximizer is given by $\textsf{id}$. For general $k$, to the best of
our knowledge, there is no efficient algorithm for computing the maximizer directly. However, there exists a Lagrangian multiplier $\gamma > 0$ such that \eqref{eq:sparsity_constrained} is equivalent to the optimization problem 
\begin{equation}\label{eq:sparsity_LAP}
\max_{\pi \in \mc{P}(n)} \left\{ \su x_{\pi(i)} y_i - \gamma \dH(\pi, \textsf{id}) \right \}  = \max_{\Pi \in \mc{P}(n)} \left\{ \su \sum_{j = 1}^n \pi_{ij} (x_j y_i - \gamma \mathbb{I}(i \neq j))  \right \},
\end{equation}
which is a linear assignment problem with cost matrix $C = \big(\gamma  \mathbb{I}(i \neq j) - x_j y_i \big)_{i,j}$, which is computationally tractable according to the discussion following \eqref{eq:LAP} below. As elaborated in the next subsection $\S$\ref{subsec:expfam}, the maximizer of \eqref{eq:sparsity_LAP} corresponds to the MAP estimator of $\pi^*$ (for fixed model parameter $\beta$) under
a specific class of prior distributions over $\mc{P}(n)$. The right panel of Figure \ref{fig:motivation_sparse} highlights the improvement that can be achieved by the resulting estimator which here only performs a small number
of re-pairings capturing those pairs that correspond to massive mismatch error in the left panel. This re-pairing is consistent with the underlying regression slope, avoiding both the amplification bias associated
with $\wh{\pi}_{\text{ML}}$ depicted in the middle panel as well as the attenuation bias that is incurred when ignoring mismatch error altogether (e.g., \cite{Neter65, Scheuren93, Scheuren97, WangBenDavidDiaoSlawski2021}) . 

\begin{figure}
   \includegraphics[height = .15\textheight]{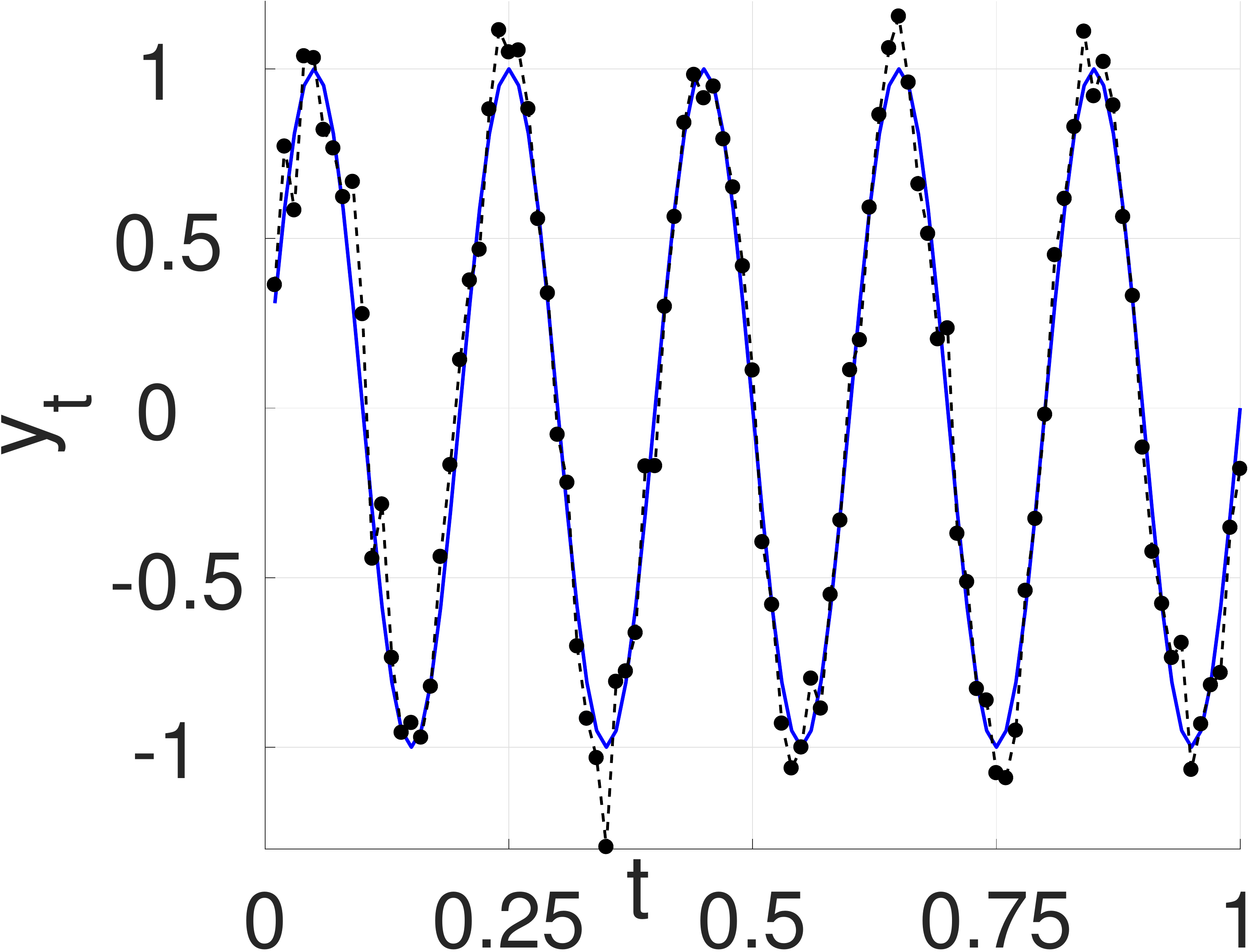}  \hspace*{1ex} \includegraphics[height = .15\textheight]{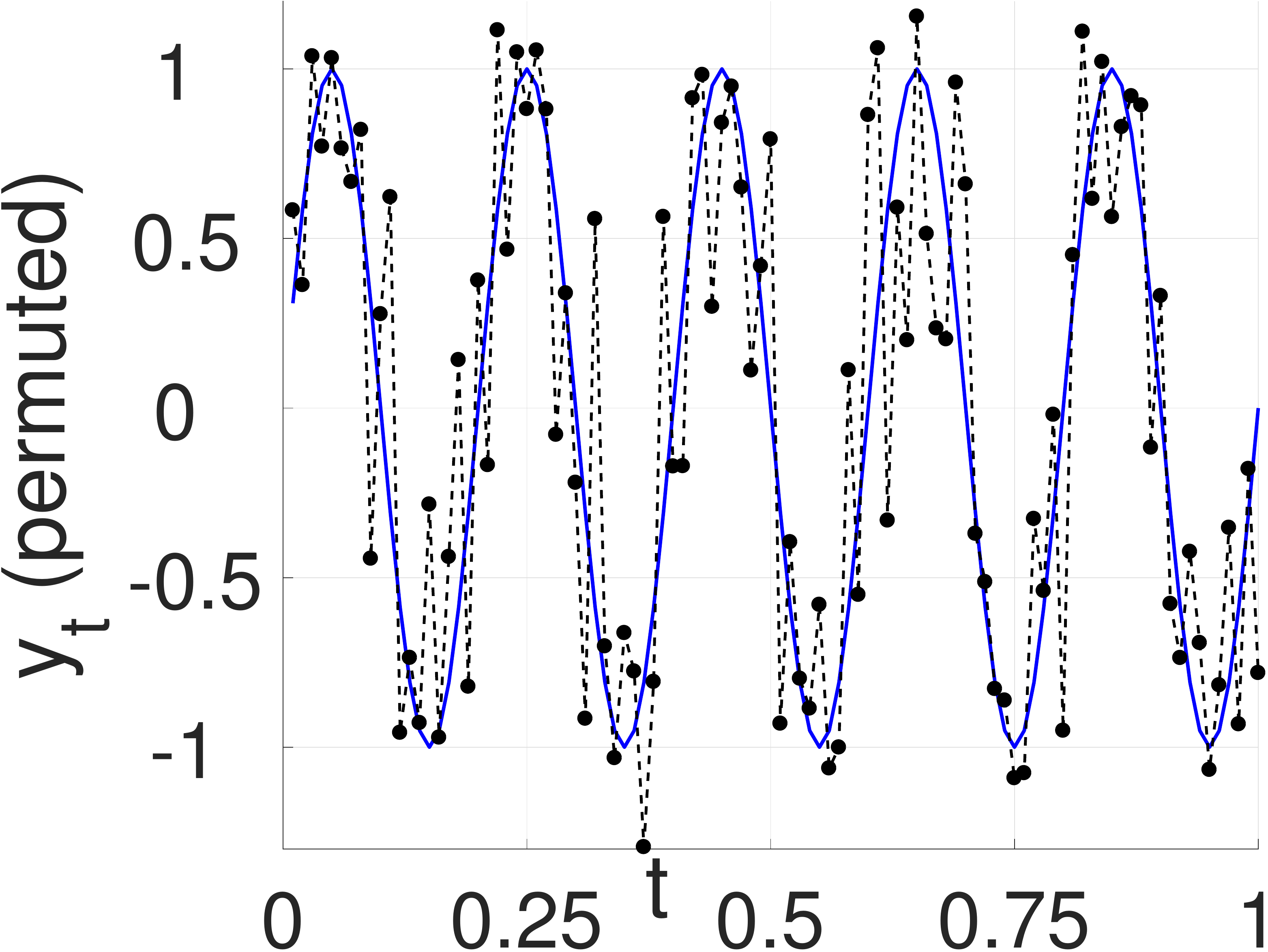} \hspace*{1ex} \includegraphics[height = .15\textheight]{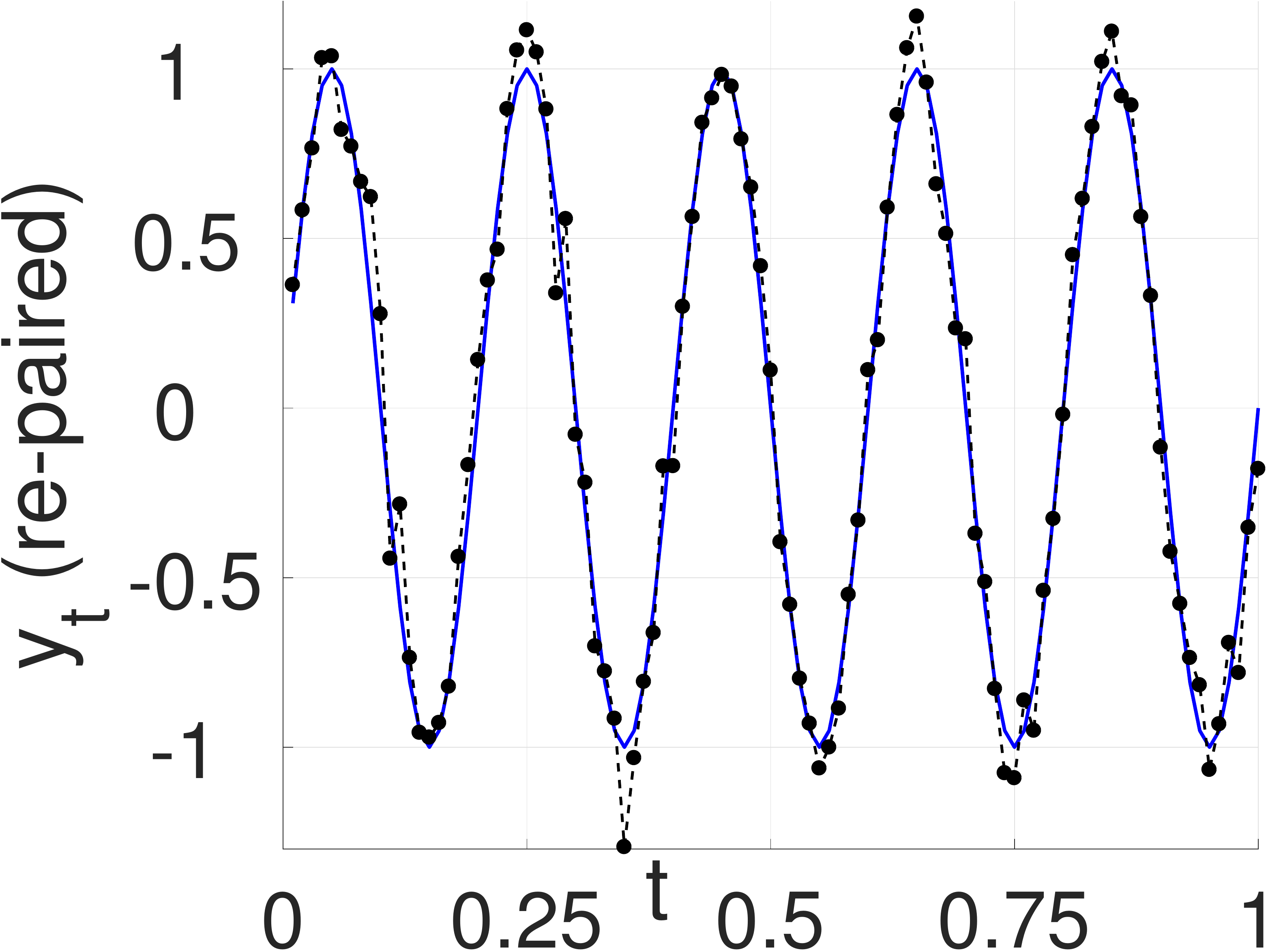}
   \vspace*{-2ex}
    \caption{L: Original data $y_{t_i} = \sin(10 \pi t_i) + 0.1 \eps_i$, $\eps_i \sim N(0,1)$, $1 \leq i \leq n$.  M: Locally permuted data $(t_i, y_{t_{\pi^*(i)}})$ R: Corrected data $(t_i, y_{t_{\wh{\pi} \circ \pi^*(i)}})$ based on the prior discussed in the text.} \label{fig:motivation_local}
\end{figure}

Figure \ref{fig:motivation_local} illustrates scenarios in which $\pi^*$ is not sparse (with a mismatch rate 
exceeding 80\%), but constrained to be a ``local shuffling" in the sense that $\max_{i \in [n]} |\pi^*(i) - i| \leq r$, i.e., the corresponding permutation matrix is a band matrix with bandwidth at most $r$. This scenario
is particularly relevant when the data is recorded sequentially (e.g., over different time points) or across a spatial domain endowed with a notion of distance, and it is known that $\pi^*$ can only mix up the order of 
data inside a specific time window or within a local neighborhood. There are numerous applications in which $\pi^*$ is locally constrained such as genome sequencing \cite{Abid2017}, signal processing \cite{Balakrishnan1962, Abbasi2021}, or computer vision \cite{Ma2021}. 

The illustrative example in Figure \ref{fig:motivation_local} can be thought of as a regression problem in which the signal is  a sine with known frequency but unknown (positive) amplitude $\beta^*$, i.e., $y_{t_i} = \beta^* \sin(10 \pi t_i) + 0.1 \eps_i$, $1 \leq i \leq n$ (left panel). However, the observed data is of the form $(y_{t_{\pi^*(i)}})_{i = 1}^n$ for some unknown (local) permutation $\pi^*$ (middle panel). If $\beta^*$ is known to be positive, then the (unconstrained) ML estimator  $\wh{\pi}_{\text{ML}}$ of $\pi^*$ matches the order statistics $\{ \mu_{(i)} \}_{i = 1}^n$ and $\{ y_{(i)} \}_{i = 1}^n$, where $\mu_i = \sin(10 \pi t_i)$, $1 \leq i \leq n$. In order to improve over the ML estimator using the prior knowledge of local shuffling, we impose the constraint that the alternative estimator $\wh{\pi}$ does not pair any indices that are more than $r = 3$ apart. This estimator can be obtained as solution of the optimization problem 
\begin{equation}\label{eq:local_LAP}
\max_{\substack{\pi \in \mc{P}(n) \\
\max_{i \in [n]}|\pi(i) - i| \leq r}} \left\{ \su \mu_i y_{\pi(i)}  \right \}  = \max_{\Pi \in \mc{P}(n)} \left\{ \su \sum_{j = 1}^n \pi_{ij} (\mu_i y_j - c_{ij})  \right \},
\end{equation}
where $c_{ij} = 0$ if $|i - j| \leq r$ and $c_{ij} = +\infty$ otherwise. As in \eqref{eq:sparsity_LAP}, the problem on the right hand side is a linear assignment problem and hence computationally tractable, and corresponds to MAP
estimation under the family of priors considered below in detail. The corrected, i.e., repaired data 
$(t_i, y_{\wh{\pi} \circ \pi^*(i)})_{i = 1}^n$ based on this approach are depicted in the right panel of Figure \ref{fig:motivation_local}. 

\subsection{Exponential family prior on $\mc{P}(n)$}\label{subsec:expfam}
In this subsection, we show that the priors discussed in the two examples of the previous subsection can be understood
as specific instances of a more general family of prior distributions over $\mc{P}(n)$. Specifically, we consider
the family of priors
\begin{equation}\label{eq:prior}
p(\pi) \propto \exp(\gamma \, \tr(\Pi^{\T} M)), \quad M \in \R^{n \times n}, \quad \gamma > 0, 
\end{equation}
where $\gamma > 0$ is the concentration parameter, and the matrix $M$ (which is not required to have any specific properties)  
defines the mode(s) of the distribution, i.e., $\argmax_{\Pi \in \mc{P}(n)} \nscp{\Pi}{M}$, where $\nscp{\cdot}{\cdot}$ here
represents the trace inner product on matrices of the same dimension that induces the Frobenius norm $\nnorm{\cdot}_{\textsf{F}}$. In the same vein, the mode(s)
of the distribution correspond to the set of matrices closest to $M$ with respect to the same norm. Moreover, the distribution
specified by \eqref{eq:prior} is of exponential family form with respect to the trace inner product (cf., e.g., $\S$3.2 in \cite{wainwright2008graphical}). 
\vskip1ex
\noindent \emph{Linear Assignment Problems}. Linear Assignment problems (LAPs) are a well-studied class of optimization problems
for computing optimal one-to-one matchings of two sets of items \cite{Burkard2009}. LAPs are of the form
\begin{equation}\label{eq:LAP}
\min_{\Pi \in \mc{P}(n)} \nscp{\Pi}{C},    
\end{equation}
where $C$ is a given cost matrix. By the Birkhoff-von Neumann theorem \cite{Ziegler1995}, the vertices of the set of $n$-by-$n$ doubly stochastic matrices
\begin{equation*}
\mc{D}\mc{S}(n) = \left\{P \in \R^{n \times n}: \; P_{ij} \geq 0, \; 1 \leq i,j \leq n, \;  \sum_{j= 1}^n P_{ij} = 1, \; \, 1 \leq i \leq n, \; \sum_{i= 1}^n P_{ij} = 1, \; \, 1 \leq j \leq n \right \}    
\end{equation*}    
are given by $\mc{P}(n)$. As a result, the minimum over $\mc{P}(n)$ can be replaced by the minimum over $\mc{D}\mc{S}(n)$, hence
\eqref{eq:LAP} reduces to a linear program in $n^2$ variables and $n^2 + 2n$ linear constraints. 

This brief summary entails that computing a mode of  \eqref{eq:prior} reduces to a tractable problem, by setting $C = -M$ in \eqref{eq:LAP}.

\vskip1ex
\noindent \emph{Specific examples}. Below, we consider a few examples of interest that are special cases of
\eqref{eq:prior}. 
\vskip1ex
\noindent (I) Hamming prior. \\
\noindent Consider the choice $M = I_n$. In this case, for any $\Pi \in \mc{P}(n)$, we have 
\begin{equation*}
\nscp{\Pi}{I_n} = \sum_{i = 1}^n \Pi_{ii} = n - \sum_{i = 1}^n \mathbb{I}(\Pi_{ii} \neq 1) = n - \dH(\pi, \textsf{id}), 
\end{equation*}
where, as before, $\dH(\pi, \pi') = \sum_{i = 1}^n \mathbb{I}(\pi(i) \neq \pi'(i))$ denotes the Hamming distance on $\mc{P}(n)$. 
Since $n$ does not depend on $\pi$, this implies that \eqref{eq:prior} can be expressed equivalently as 
\begin{equation}\label{eq:Hammingprior}
p(\pi) \propto \exp(-\gamma \, \dH(\pi, \textsf{id})),
\end{equation}
which appeared in the first example of the preceding subsection $\S$\ref{subsec:mot_example}, cf.~\eqref{eq:sparsity_LAP}, in which the goal
was to take into account the underlying low rate of mismatches. The distribution \eqref{eq:Hammingprior} is 
a specific instance of the class of Mallow's priors of the form $p(\pi) \propto \exp(-\gamma d(\pi, \pi_0))$
for a base permutation $\pi_0$ and a metric $d$ on $\mc{P}(n)$ \cite{Mallows1957, Diaconis1988, fligner1986distance, irurozki2019mallows, fligner1993probability}. We note, however, that the family of Mallow's priors is not a sub-family
of the exponential family prior \eqref{eq:prior} since the metric $d$ cannot be expressed via a trace inner product
in general. 
\vskip1ex
\noindent (II) Local shuffling prior. \\
As in the second example in $\S$\ref{subsec:mot_example}, suppose we want to have the prior $p$ place most of its
mass on permutations that move indices within small windows, i.e., $|\pi(i) - i|$ tends to be small. This can be achieved by choosing the entries of the matrix $M$ in \eqref{eq:prior} of the form $M_{ij} = -\phi(|i - j|)$ for some non-decreasing function $\phi: \R_+ \rightarrow \R_+$. The choice $\phi(u) = 0$ if $u \leq r$ for some positive integer $r$ and $\phi(u) = +\infty$ otherwise 
yields the approach \eqref{eq:local_LAP} that underlies the example in Figure \ref{fig:motivation_local} above. 
\vskip1ex
\noindent (III) Block prior. \\
Under the local shuffling assumption $\max_{i \in [n]} |\pi^*(i) - i| \leq r$, the corresponding permutation matrix $\Pi^*$ is typically block diagonal, i.e., $\Pi^* = \text{bdiag}(\Pi_1^*, \ldots, \Pi^*_B)$,
with all $B$ blocks having size proportional to $r$. Note that the composition of the blocks is generally \emph{not} known in advance. On the other hand, in record linkage applications, it is rather common that the composition of the blocks is indeed known given matching variables used during record linkage. For example, suppose that the combination of gender, ethnicity, and age group are used for that purpose and that these three categorical variables are free of errors. In this case, mismatches can only involve pairs $(i,j)$ falling into the same block corresponding to a specific combination of gender, ethnicity, and age. Such known block structure can be encoded via prior \eqref{eq:prior} by choosing $M_{ij} = -\infty$ if $(i,j)$ is not contained in the same block and $M_{ij} = 0$ otherwise. Note that this corresponds to a uniform prior for each block, i.e., $p(\pi) = \prod_{b = 1}^B p(\pi_b)$ with $p(\pi_b) \propto 1$, $1 \leq b \leq B$. The prior for each block does not have to be necessarily uniform. For example, a Hamming prior as in Example (I) above can be used instead. Moreover, the hard block constraint can be softened by choosing a suitable finite negative number for entries $M_{ij}$ corresponding to pairs
$(i,j)$ not contained in the same block. 
\vskip1ex
\noindent (IV) Lahiri-Larsen prior. \\
In their seminal work on adjusting (generalized) linear regression in the presence of mismatch errors, Lahiri \& Larsen \cite{Lahiri05} and Chambers \cite{Chambers2009} assume that $\pi^* \sim p(\pi)$ whose expectation $\E_{p(\pi)}[\Pi^*] = Q$   
is known to the (post-linkage) data analyst. Observe that $Q \in \mc{D}\mc{S}(n)$ is a double stochastic matrix. Any distribution over permutations whose expectation equals $Q$ can be used as a potential prior. In fact, it is known that there are distributions supported over only $O(n^2)$ (as opposed to $n!$) permutations satisfying that requirement. Identification
of the support via Birkhoff's decomposition \cite{johnson_dulmage_mendelsohn_1960} requires at least $O(n^4)$ runtime, and hence is not scalable. In the framework considered here, it is convenient to use $M = Q$ in the prior \eqref{eq:LAP}. The mode(s) of that prior are then given by the permutation matrices solving the Euclidean projection problem $\min_{\Pi \in \mc{P}(n)} \nnorm{\Pi - Q}_{\textsf{F}}^2$ of $Q$ on $\mc{P}(n)$. A basic example for $Q$ results for the so-called exchangeable linkage model in \cite{Chambers2009, Zhang2020} in which $Q = (1 - \alpha) I_n + \frac{\alpha}{n-1} \bm{1}_n \bm{1}_n^{\T}$. Since in this case $\scp{Q}{\Pi} = (1 - \alpha)\scp{I_n}{\Pi} + \alpha \frac{n}{n-1}$ for $\alpha \in [0,1)$, the resulting prior is equivalent to the Hamming prior considered in Example (I). More complex priors are obtained depending on the structure of $Q$. 

\subsection{Integration in Shuffled Data Problems}\label{subsec:dataproblems}
In this subsection, we outline how the above prior can be integrated into generic shuffled data problems. The proposed 
Monte-Carlo EM \cite{wei1990monte} framework builds upon the classical work \cite{Wu1998} that has been rediscovered in the more recent work \cite{Abid2018}. The Monte-Carlo EM scheme in \cite{Wu1998} was further developed based on the concept
of data augmentation \cite{TannerWong1987} in Gutman \emph{et al}.~\cite{Gutman13}. Note that none of \cite{Wu1998, Abid2018, Gutman13} consider informative priors for the permutation. 
\vskip2ex
\noindent {\bfseries Conditional and Integrated Likelihood}
\vskip1ex
\noindent 
Suppose we are given data $\mc{D} = \{ (\M{x}_i, \M{y}_i)\}_{i = 1}^n$ potentially contaminated by mismatch error. Let $p(\M{x}_j, \M{y}_i;\theta)$ be the likelihood (depending on a parameter $\theta$) for the pairing of $\M{x}_j$ with $\M{y}_i$, $(i,j) \in [n]^2$. The likelihood for $\theta$ resulting from $\mc{D}$ conditional on a specific permutation $\pi \in \mc{P}(n)$ is given by 
\begin{equation}\label{eq:conditional_likelihood}
L(\theta| \pi) = \prod_{i = 1}^n p(\M{x}_{\pi(i)}, \M{y}_i;\theta) = \prod_{i = 1}^n \prod_{j = 1}^n p(\M{x}_j, \M{y}_i;\theta)^{\pi_{ij}}
\end{equation}
\emph{Conjugacy}. It is worth noting that under \eqref{eq:conditional_likelihood}, the posterior $p(\pi|\mc{D}, \theta)$ is a member of the 
family of distributions specified by $p(\pi)$ of the form \eqref{eq:prior}, i.e., the latter is a conjugate prior. This follows immediately from the observation that 
\begin{align}
p(\pi|\mc{D}, \theta) \propto p(\mc{D}|\pi, \theta) \cdot p(\pi) &= L(\theta|\pi) \cdot p(\pi) \notag \\
&= \exp\left(\sum_{i = 1}^n \sum_{j = 1}^n \pi_{ij} \left[ \log\{ p (\M{x}_j, \M{y}_i | \theta) \} + \gamma M_{ij}  \right] \right) \notag\\
&= \exp(\tr(\Pi^{\T} M_{\mc{D}, \theta, \gamma})),\label{eq:proof_conjugacy} 
\end{align}
with $M_{\mc{D}, \theta, \gamma} = \big( \log(p(\M{x}_j, \M{y}_i | \theta)) + \gamma M_{ij} \big)$. 

The (conditional) likelihood \eqref{eq:conditional_likelihood} can be maximized with respect to both $\theta$ and $\pi$ as, e.g.,  in \cite{Pananjady2016, Abid2017, SlawskiBenDavid2017}. In an alternative view, $\theta$ is considered as the quantity of primary interest, in which case one would rather consider the \emph{integrated likelihood} 
\begin{equation}\label{eq:integrated_likelihood}
L(\theta) = \E_{\pi} [ L(\theta | \pi)] = \sum_{\pi \in \mc{P}(n)} L(\theta|\pi) p(\pi). 
\end{equation}
As seen in $\S$\ref{subsec:mot_example}, maximizing the conditional likelihood tends to be prone to overfitting, prompting a need for regularization. The use of the integrated likelihood mitigates that problem at best slightly, but not substantially as can be seen, e.g., by examining the case of linear regression with i.i.d.~Gaussian errors (cf.~Appendix \ref{app:overfitting_integrated}), hence regularization remains relevant.
\vskip1ex
\noindent {\bfseries Monte Carlo EM scheme}
\vskip1ex
\noindent 
The Expectation-Maximization (EM) algorithm \cite{dempster1977maximum} is an established heuristic for (approximately) minimizing the negative log-likelihood corresponding to \eqref{eq:integrated_likelihood}. Specifically, instead of minimizing $\ell(\theta) = -\log L(\theta)$, a sequence of surrogate functions $\{ \wt{\ell}^{(t)}(\theta; \theta^{(t)}) \}_{t \geq 0}$ are minimized successively: 
\begin{equation*}
\ell(\theta) = -\log \E_{\pi}[L(\theta|\pi)] \; \; \leadsto \; \; \E_{\pi|\mc{D},\theta^{(t)}}[-\log L(\theta|\pi)] \invcoloneq \wt{\ell}^{(t)}(\theta; \theta^{(t)}), 
\end{equation*}
where  
\begin{equation}\label{eq:expected_complete}
\E_{\pi|\mc{D},\theta^{(t)}}[-\log L(\theta|\pi)] = \sum_{i = 1}^n \sum_{j = 1}^n \E[\pi_{ij} | \mc{D}, \theta^{(t)}] \{ -\log p(\M{x}_j, \M{y}_i; \theta) \}   
\end{equation}    
is the so-called \emph{expected complete data} negative log-likelihood. The surrogates $\{ \wt{\ell}^{(t)}(\cdot\,;\theta^{(t)}) \}$ tend to be easier to minimize since they are linear combinations of standard likelihood terms as they are encountered for fixed and known $\pi$. Surrogates are updated according to the recursion 
\begin{equation*}
\theta^{(t+1)} \leftarrow \argmin_{\theta} \wt{\ell}^{(t)}(\theta; \theta^{(t)}) \; \; \leadsto \; \; \wt{\ell}^{(t+1)}(\theta;\theta^{(t+1)}).     
\end{equation*}    
It can be shown \cite{dempster1977maximum} that the sequence generated in this fashion exhibits a monotonic descent property with respect to the original objective $\ell$, i.e., $\ell(\theta^{(t+1)}) \leq \ell(\theta^{(t)})$ for all $t$. 

Here, the main challenge of this scheme is the E-step, i.e, the calculation of the expectation on the right term in \eqref{eq:expected_complete}. For any entry $\pi_{ij}$ of $\Pi$, we have 
\begin{align*}
\E[\pi_{ij}|\mc{D}, \theta^{(t)}] = \sum_{\pi \in \mc{P}(n)} p(\pi|\mc{D},\theta^{(t)}) \pi_{ij} &\propto \sum_{\pi \in \mc{P}(n)}  p(\mc{D}|\pi,\theta^{(t)}) p(\pi) \pi_{ij} \\
&= \sum_{\pi \in \mc{P}(n)} \left\{ \prod_{i = 1}^n p(\M{x}_{\pi(i)}, \M{y}_i;\theta^{(t)}) \right \}p(\pi) \pi_{ij}. 
\end{align*}    
Since the summation over $\mc{P}(n)$ is not computationally tractable, the expectation needs to be approximated, e.g., via
Monte Carlo simulation. Since for the same reason, the posterior $p(\pi | \mc{D}, \theta^{(t)})$ is only accessible up to an unknown constant (cf.~\eqref{eq:proof_conjugacy}), it is appropriate to resort to Markov Chain Monte Carlo (MCMC) sampling techniques \cite{gelman2013bayesian}. The Metropolis-Hastings (MH)
algorithm can be used to generate a Markov Chain $\{ \pi^{(k)} \}_{k \geq 1}$ whose stationary distribution equals $p(\pi|\mc{D}, \theta^{(t)})$. This yields the approximation
\begin{equation}\label{eq:MC-EM}
\wh{\E}[\pi_{ij}|\mc{D}, \theta^{(t)}] = \frac{1}{m - b} \sum_{k = b+1}^m  \pi_{ij}^{(k)},  \quad (i,j) \in [n]^2. 
\end{equation}    
where $b$ denotes the length of the so-called ``burn-in" period after which the Markov Chain is considered to have reached stationarity, and $m$ denotes the total length of the Markov chain. Substituting \eqref{eq:MC-EM} into \eqref{eq:expected_complete} then yields what is known as Monte-Carlo EM
scheme, summarized in Algorithm \ref{alg:montecarloEM}.

Conveniently, there is a proposal distribution for the MH algorithm that is, in a sense, canonical and easy to work with.
That proposal distribution, known as Fisher-Yates sampling, generates a new permutation from the current one by swapping 
the assignments of a pair of indices (cf.~Algorithm \ref{alg:MH} for details).
\vskip2ex
\noindent \emph{Initialization}. It is a well-known fact that the choice of the initial iterate $\theta^{(0)}$ can critically impact the quality of the solution that is returned by EM schemes given that the latter is a local strategy that finds a stationary point of a (in general) non-convex objective near the initial iterate. Several consistent initial estimators of $\theta^*$ are known for regression setups depending on the structure of $\pi$ \cite{SlawskiBenDavid2017, SlawskiDiaoBenDavid2019, Lahiri05, Chambers2009, chambers2019improved, Peng2021}, and those naturally lend themselves as initial iterate for the above EM scheme.

Careful initialization of the MH subroutine is important as well in order to ensure that $p(\pi|\mc{D}, \theta^{(t)})$ is explored reasonably well given that the domain of interest $\mc{P(n)}$ has $n!$ elements while the number of MCMC iterations $m$ is limited. Fortunately, under the prior \eqref{eq:prior}, computing the mode $\argmax_{\pi} p(\pi|\mc{D}, \theta^{(t)})$ reduces to an LAP of the form \eqref{eq:LAP} in virtue of \eqref{eq:proof_conjugacy}. Initialization via the mode has the advantage that the Markov chain is started in a high density region. The hope is that the resulting iterates (which are generated according to a localized proposal distribution) will pick up most of the mass of $p(\pi|\mc{D}, \theta^{(t)})$ so that \eqref{eq:MC-EM} will well approximate the underlying expectation.   

\vskip1ex
\begin{minipage}{\textwidth}
\hspace*{-3.5ex}
\begin{minipage}{0.63\textwidth}
\vspace*{-23.5ex}
\begin{algorithm}[H]
\caption{Monte Carlo EM algorithm}\label{alg:montecarloEM}
{\bfseries Input}: $\mc{D} = \{\{ \M{x}_i \}_{i = 1}^n$,  $\{ \M{y}_i \}_{i = 1}^n \}$, $\gamma$, \verb+EM_iter+  \par 
\textbf{Initialize} $\theta^{(0)} \leftarrow \wh{\theta}_{\text{init}}$. \par
{\bfseries for} $t=0,\ldots,$\,\verb+EM_iter+
\begin{itemize}[leftmargin = .85ex]
\item[] $\wh{\pi}_{\text{init}} \leftarrow \argmax_{\pi \in \mc{P}(n)} p(\pi | \mc{D}, \theta^{(t)})$.
\item[] $\wh{\E}[\pi|\mc{D}, \theta^{(t)}] \leftarrow \texttt{MH}(\mc{D}, \theta^{(t)}, \wh{\pi}_{\text{init}}, \gamma, m)$.
\item[] $\theta^{(t+1)} \leftarrow \min_{\theta} \mbox{{\small $\left\{ \sum_{i = 1}^n \sum_{j = 1}^n \wh{\E}[\pi_{ij} | \mc{D}, \theta^{(t)}] \{ -\log p(\M{x}_j, \M{y}_i; \theta) \} \right\}$}}$.  
\item[] $t \leftarrow t+1$
\end{itemize}
{\bfseries end for}
\end{algorithm}
\end{minipage}
\begin{minipage}{0.36\textwidth}
\begin{algorithm}[H]
\caption{\texttt{MH} sub-routine}\label{alg:MH}
{\bfseries Input}: $\mc{D},  \theta, \wh{\pi}_{\text{init}}, \gamma, m$\par 
\textbf{Initialize} $\pi^{(0)} \leftarrow \wh{\pi}_{\text{init}}$. \par
{\bfseries for} $k=0,\ldots,m$
\begin{itemize}[leftmargin = .85ex]
\item[] Sample $(i,j) \in [n]^2$\footnote{Uniformly at random, or based on more sophisticated schemes.}.
\item[] $\wt{\pi}(i) \leftarrow \pi^{(k)}(j)$, $\wt{\pi}(j) = \pi^{(k)}(i)$. 
\item[] $r(\wt{\pi}, \pi^{(k)}) \leftarrow \min \left\{ \frac{p(\wt{\pi}|\mc{D}, \theta;\gamma)}{p(\pi^{(k)}|\mc{D}, \theta;\gamma)}, 1  \right \}$.
\item[] Draw $u \sim U([0,1])$.
\item[] {\bf if} $r(\wt{\pi}, \pi^{(k)}) > u$: $\pi^{(k + 1)} \leftarrow \wt{\pi}$.
\item[] {\bf else}: $\pi^{(k + 1)} \leftarrow \pi^{(k)}$.
\item[] $k \leftarrow k+1$. 
\end{itemize}
{\bfseries end for} \\
{\bfseries return} $\wh{\E}[\pi | \mc{D}, \theta]$ as in \eqref{eq:MC-EM}
\end{algorithm}
\end{minipage}
\end{minipage}

\vskip2ex
\noindent {\bfseries Reduction under exponential family likelihood}
\vskip1ex
\noindent Interestingly, for a variety of exponential family models, the expected complete data negative log-likelihood \eqref{eq:expected_complete}
involves only $n$ instead of $n^2$ terms, which is a substantial reduction. Specifically, \eqref{eq:expected_complete}
will be of the form 
\begin{equation*}
\sum_{i = 1}^n r\{\M{x}_i, \M{y}_i, (\E[\Pi | \mc{D}, \theta^{(t)}]^{\T} \M{Y})_i; \theta) \}
\end{equation*}
for a function $r$ depending at most on $\{ \M{x}_i, \M{y}_i, (\E[\Pi | \mc{D}, \theta^{(t)}]^{\T} \M{Y})_i \}_{i = 1}^n$. Specific examples of interest are presented in the sequel.\vskip1ex 
\noindent \emph{(i) Least squares regression}. Taking $-\log p(\M{x}, y; \beta, \sigma^2) = \frac{1}{2\sigma^2} (y - \M{x}^{\T} \beta)^2$, which corresponds to the negative likelihood of a linear regression model with i.i.d.~zero-mean Gaussian errors with variance $\sigma^2$ yields the following expression for the expected complete data negative log-likelihood:
\begin{align*}
&\frac{1}{2\sigma^2} \sum_{i = 1}^n \sum_{j = 1}^n \E[\pi_{ij} | \mc{D}, \theta^{(t)}] (y_i - \M{x}_j^{\T} \beta)^2 
= \frac{1}{\sigma^2} \sum_{i = 1}^n \sum_{j = 1}^n \E[\pi_{ij} | \mc{D}, \theta^{(t)}] \left\{ \frac{1}{2} (\M{x}_j^{\T} \beta)^2 - y_i \M{x}_j^{\T} \beta \right \} + c.
\\
&= \frac{1}{2\sigma^2} \sum_{i = 1}^n \E \left[\sum_{j = 1}^n \pi_{ij} (\M{x}_j^{\T} \beta)^2 \Big | \mc{D}, \theta^{(t)} \right]    - \frac{1}{\sigma^2} \sum_{j = 1}^n \M{x}_j^{\T} \beta \sum_{i = 1}^n \E[\pi_{ij} | \mc{D}, \theta^{(t)}] y_i + c \\
&= \frac{1}{\sigma^2} \left\{ \frac{1}{2} \sum_{i = 1}^n (\M{x}_i^{\T} \beta)^2 - \sum_{i = 1}^n \left \{ \E[\Pi | \mc{D}, \theta^{(t)}]^{\T} \M{Y} \right\}_i (\M{x}_i^{\T} \beta) \right\} + c \\
&=\frac{1}{\sigma^2} \left\{ \frac{1}{2} \nnorm{\M{X} \beta}_2^2 - \nscp{\E[\Pi | \mc{D}, \theta^{(t)}]^{\T} \M{Y}}{\M{X} \beta} \right\} + c
\end{align*}
%
which is identical to a standard least squares objective with design matrix  $\M{X}$ and response vector $\E[\Pi^{\T} | \mc{D}, \theta^{(t)}] \M{Y}$.   
\vskip1ex
\noindent \emph{(ii) Generalized linear models}. In this case, we have $-\log p(\M{x}, y;\beta, \phi) = \frac{\psi(\M{x}^{\T} \beta) - y \M{x}^{\T} \beta}{a(\phi)} + c(y, \phi)$, where $a$, $\psi$ and $c$ denote scale function, cumulant, and partition function, respectively. Using a reasoning similar to above, one shows that 
\begin{align*}
&\frac{1}{a(\phi)}\sum_{i = 1}^n \sum_{j = 1}^n \E[\pi_{ij} | \mc{D}, \theta^{(t)}] \{ \psi(\M{x}_j^{\T} \beta) - y_i \M{x}_j^{\T} \beta \} + \sum_{i = 1}^n c(y_i, \phi) \\
=&\frac{1}{a(\phi)} \su \{ \psi(\M{x}_i^{\T} \beta) - ( \E[\Pi | \mc{D}, \theta^{(t)}]^{\T} \M{Y})_i \, \M{x}_i^{\T} \beta \} + \sum_{i = 1}^n c(y_i, \phi). 
\end{align*}
While in the above calculation, the canonical link is assumed for simplicity, this assumption is not necessary to achieve the aforementioned reduction
from $n^2$ to $n$ terms. 
\vskip1ex
\noindent \emph{(iii) Multivariate Normal data and precision matrix estimation}. Suppose that $(\M{x}, \M{y}) \sim N(\mu_*, \Omega_*^{-1})$, where $\Omega_*$ is referred to as precision matrix. Since estimation of $\mu_*$ is not affected by the presence
of an unknown permutation, let us assume that $\mu_* = 0$ for simplicity. We have $-\log p(\M{x}, \M{y}; \Omega) = -\log \det \Omega + \tr(\Omega \M{z} \M{z}^{\T})$ up to additive constants, where $\M{z} = [\M{x}^{\T} \; \M{y}^{\T}]^{\T}$ denotes the horizontal concatenation of $\M{x}$ and $\M{y}$. Note that $\M{z} \M{z}^{\T}$ consists of diagonal blocks $\M{x} \M{x}^{\T}$ and 
$\M{y} \M{y}^{\T}$ and off-diagonal blocks $\M{x} \M{y}^{\T}$ and $\M{y} \M{x}^{\T}$. Accordingly, we have $\tr(\Omega \M{z} \M{z}^{\T}) = \tr(\Omega_{\M{x}\M{x}} \M{x} \M{x}^{\T}) + \tr(\Omega_{\M{y}\M{y}} \M{y} \M{y}^{\T}) + 2 \tr(\Omega_{\M{y} \M{x}} \M{x} \M{y}^{\T})$, where $\Omega_{\M{x} \M{x}}$, $\Omega_{\M{y} \M{y}}$ etc.~denote the corresponding sub-matrices of $\Omega$. Consequently, we have (up to constants)
\begin{align*}
&\su \sum_{j = 1}^n \E[\pi_{ij} | \mc{D}, \Omega^{(t)}] \{-\log p(\M{x}_j, \M{y}_i; \Omega) \} \\\
&= -n \log \det \Omega + \tr\left(\Omega_{\M{x} \M{x}} \su \M{x}_i \M{x}_i^{\T} \right) + \tr\left(\Omega_{\M{y} \M{y}} \su \M{y}_i \M{y}_i^{\T} \right) + \tr \left(\Omega_{\M{y} \M{x}} \su \sum_{j = 1}^n  \E[\pi_{ij} | \mc{D}, \Omega^{(t)}] \M{x}_j \M{y}_i^{\T}  \right) \\
&= -n \log \det \Omega + \tr(\Omega_{\M{x} \M{x}} \M{X}^{\T} \M{X}) + \tr(\Omega_{\M{y} \M{y}} \M{Y}^{\T} \M{Y}) +  \tr(\Omega_{\M{y} \M{x}} \M{X}^{\T} \E[\Pi|\mc{D}, \Omega^{(t)}]^{\T} \M{Y})\\
&=  -n ( \log \det \Omega + \tr(\Omega \M{S}_{ \E[\Pi|\mc{D}, \Omega^{(t)}]}),
\end{align*}
where $\M{S}_{ \E[\Pi|\mc{D}, \Omega^{(t)}]}$ consists of blocks {\small $\M{X}^{\T} \M{X} / n$, $\M{Y}^{\T} \M{Y} / n$, $\M{X}^{\T} \E[\Pi|\mc{D}, \Omega^{(t)}]^{\T} \M{Y}/n$ and $\M{Y}^{\T} \E[\Pi|\mc{D}, \Omega^{(t)}] \M{X}/n$}.  
\vskip2ex
\noindent {\bfseries Computational complexity}
\vskip1ex
\noindent We here discuss the computational complexity of Algorithm \ref{alg:montecarloEM}. For exponential family models benefiting from the above reduction, the M-step, i.e., the update of $\theta$ only involves $n$ terms, and is hence computationally equivalent to a standard estimation problem. Apart from the initialization of the Markov chain, the approximate E-step has complexity $O(m)$, where $m$ denotes the length of the Markov chain. Computing the acceptance probability, updating $\pi^{(k)}$, and keeping tracking of $\wh{\E}[\Pi | \mc{D}, \theta^{(t)}]^{\T} \M{Y}$ within Algorithm \ref{alg:MH} can be done in time $O(1)$ given that the proposal distribution only changes $\pi^{(k)}$ at two positions. However, $m$ is recommended to be at least of the order $\Omega(n)$, heuristically justified by the fact that in the worst case a permutation is the product of $n-1$ transpositions. Obtaining the initial permutation $\wh{\pi}_{\text{init}}$ generally involves the solution of a linear assignment problem, which is costly with a complexity of $O(n^3)$, while approximate solutions via Sinkhorn iterations can be obtained in complexity $O(n^2)$ \cite{COT2019}. Alternatively, since computing $\wh{\pi}_{\text{init}}$ based on an LAP is expensive, $\wh{\pi}_{\text{init}}$ can be initialized as the last state of the Markov chain returned by Algorithm \ref{alg:MH} in the previous EM iteration $t$, for all EM iterations
beyond the first.
\vskip2ex
\noindent {\bfseries Data augmentation}
\vskip1ex
\noindent Following \cite{TannerWong1987} and \cite{Gutman13}, the Monte-Carlo EM approach can be converted into a Bayesian
inference procedure targeting the posterior $p(\theta | \mc{D})$, and along the way also $p(\pi | \mc{D})$. Since
Monte-Carlo EM already involves sampling from the distribution of $p(\pi | \mc{D}, \theta)$, one can as well sample
from $p(\theta | \mc{D}, \pi)$ in an alternating fashion, which can be understood as a specific Gibbs sampler for the joint
posterior $p(\theta, \pi | \mc{D})$. Such sampling scheme is particularly attractive whenever it is easy to sample from $p(\theta | \mc{D}, \pi)$. This is the case, e.g., for the Gaussian linear regression setting with conjugate priors for the regression parameter $\beta$ and the error variance $\sigma^2$, as well as for the precision matrix estimation problem for multivariate Normal data described above with an inverse Wishart prior for $\Omega$. Compared to the Monte-Carlo EM approach, sampling from $p(\theta | \mc{D}, \pi)$ has potential advantages from the standpoint of inference (approximate standard errors and construction of credibility intervals). An illustrative example is presented in $\S$\ref{sec:experiments}.  
%

\vskip2ex
\noindent {\bfseries Beyond permutations}
\vskip1ex
\noindent We would like to point out that the framework presented herein does not require $\pi$ to be a permutation. Specifically, 
we may be given two separate files $\mc{D}_{\M{x}} = \{ \M{x}_i \}_{i = 1}^N$ and $\mc{D}_{\M{y}} = \{ \M{y}_i \}_{i = 1}^n$, $N > n$ (without loss of generality), and the consider maps $\pi: \{1,\ldots,n \}  \rightarrow \{1,\ldots,N\}$ with corresponding
matrix $\Pi \in \{0,1\}^{n \times N}$ with row sums equal to one. Letting $\mc{P}(n,N)$ denote the set of all such matrices/maps, 
we can define a prior of the form \eqref{eq:prior} given a mode $M \in \R^{n \times N}$. Conditional and integrated likelihoods
can be defined analogously to \eqref{eq:conditional_likelihood} and \eqref{eq:integrated_likelihood}, and the conjugacy property 
\eqref{eq:proof_conjugacy} continues to hold. The proposed computational Monte-Carlo EM scheme remains applicable with minor
modifications in Algorithm \ref{alg:MH}, inclusive the reduction for exponential family likelihoods.

\section{Theoretical Insights}\label{sec:theory}
In this section, we present some analysis on the proposed prior from the perspective of regularization. The main purpose
of the analysis is to provide additional guidance on the choice of the concentration parameter $\gamma$ in regression
problems. 
\vskip1ex
\noindent {\bfseries Hamming prior}. Our first results concerns the MAP estimator of $\Pi^*$ under the Hamming prior 
\eqref{eq:Hammingprior}. Specifically, we consider the linear regression setup
\begin{equation}\label{eq:gaussianlinear}
y_i = \mu_{\pi^*(i)}(\M{x}) + \sigma_* \eps_i, \quad \mu_i(\M{x}) = \M{x}_i^{\T} \beta^*, \quad \M{x}_i \sim N(0, I_d), 1 \leq i \leq n, \quad  \{ \eps_i \}_{i = 1}^n \overset{\text{i.i.d.}}{\sim}  N(0, 1),  
\end{equation}
as considered in a series of prior works on shuffled linear regression \cite{Pananjady2016, Abid2017, Hsu2017, SlawskiBenDavid2017}. The statement below considers the sparse setting with the underlying permutation $\pi^*$ satisfying the constraint $\dH(\pi^*, \textsf{id}) \leq k$ for $k$ ``small enough" as made precise below. Moreover, it is assumed for simplicity that $\beta^*$ and $\sigma_*^2$ are known; a variety of estimators for the regression parameter in this scenario have been proposed in the literature \cite{SlawskiDiaoBenDavid2019, ZhangLi2020, SlawskiBenDavid2017, Peng2021}. We recall from $\S$\ref{subsec:mot_example} that \textsf{SNR} = $\nnorm{\beta^*}_2^2 / \sigma_*^2$ denotes the signal-to-noise ratio.   
\begin{theo}\label{theo:hamming}
Suppose the setting \eqref{eq:gaussianlinear} holds true. Let $\wh{\Pi}$ denote the resulting MAP estimator of $\Pi^*$ with $d_{\emph{\textsf{H}}}(\Pi^*, I_n) \leq k$. Then, 
if $\gamma \geq 3 \gamma_0$, where $\gamma_0 = 72 \sqrt{\emph{\textsf{SNR}}} \log(en/k)$, the following holds: 
\begin{equation*}
d_{\emph{\textsf{H}}}(\wh{\Pi}, I_n) \leq 2k, \qquad \nnorm{(\wh{\Pi} - \Pi^*)\bm{\mu}}_2 \leq \sigma_* \left(17 \sqrt{k \log(e n / 3k)} + \sqrt{2 \gamma} \right).   
\end{equation*}
with probability at least $1-2/n$ and $1 -3/n$, respectively, where $\bm{\mu} = (\mu_i(\M{x}))_{i = 1}^n$. 
\end{theo}
\noindent Let us briefly comment on the implications of Theorem \ref{theo:hamming}. First, if $\gamma$ is chosen larger than
the threshold $\gamma_0$, the MAP estimator $\wh{\Pi}$ will be $2k$-sparse, which matches the sparsity of $\Pi^*$ up to the factor 
$2$. In particular, the triangle inequality implies that $\dH(\Pi^*, \wh{\Pi}) \leq 3k$, i.e., $\wh{\Pi}$ and $\Pi^*$ will be close with respect to the Hamming distance. Moreover, for values $\gamma$ such that $3\gamma_0 \leq \gamma \leq C \gamma_0$ for some
constant $C > 3$, we obtain that 
\begin{equation*}
\nnorm{(\wh{\Pi} - \Pi^*)\bm{\mu}}_2 \lesssim \sigma_* (\sqrt{k \log(en/k)} + \textsf{SNR}^{1/4}  \sqrt{k \log(en/k)}),
\end{equation*} 
where $\lesssim$ is short for $\leq$ up to constant factors. The dependence on the signal-to-noise ratio \textsf{SNR} is improved compared to the naive estimator $\wh{\Pi}_0 = I_n$,
which will scale as $\sigma_* \sqrt{k \log(en/k)} \textsf{SNR}^{1/2}$; note that for small \textsf{SNR}, one cannot
hope for improvements over $\wh{\Pi}_0$ in general. On the other hand, in light of the discussion in $\S$\ref{subsec:mot_example},
the improvement over the maximum likelihood estimator $\wh{\Pi}_{\text{ML}}$ whose corresponding error scales as $\sigma_* \sqrt{n}$, is rather substantial as long as $k$ is small relative to $n$. 
\vskip1ex
Theorem \ref{theo:hamming} provides guidance on the choice of $\gamma$ for a specific setup. To an extent, the next result yields
a lower bound on $\gamma$ in order to ensure a pre-scribed level of sparsity $k$. 
\begin{prop}\label{prop:conc_inequality} Suppose that $\pi$ follows the Hamming prior $p$ \eqref{eq:Hammingprior} with parameter $\gamma$. Then for
all $2 \leq k < n$
\begin{equation*}
\p_{\pi \sim p}\big(\text{\emph{$\dH$}}(\pi, \textsf{\emph{id}}) \geq k \big) \begin{cases} 

 \leq \exp(-k \delta \log n) \quad & \text{if} \; \gamma \geq (1 + \delta) \log n, \; \delta > 0, \\[1ex]
 \geq c(k,n)  \quad & \text{if} \; \gamma \leq \log (n-k),
 \end{cases}
\end{equation*}
where $c(k,n) \rightarrow  \frac{1}{4} \frac{!k}{k!}$ as $n \rightarrow \infty$, with $!k$ denoting the number
of derangements of $k$ elements (i.e., the number of permutations without fixed point). 
\end{prop}
\noindent Suppose it is known that $\dH(\pi^*, \textsf{id}) \leq k$. Then Proposition \ref{prop:conc_inequality} asserts
that the concentration parameter $\gamma$ of the Hamming prior \eqref{eq:Hammingprior} should be chosen proportional to
$\log(n - k) \sim \log n$ as $n$ gets large in order to ensure that the prior places essentially no mass 
outside the Hamming ball $\{\pi: \, \dH(\pi, \textsf{id}) \leq k \}$. The threshold $\gamma \sim \log n$ is sharp
in the sense that if $\gamma \leq \log(n - k)$, the prior will place at least constant mass $\frac{1}{4} !k/k! \sim \frac{1}{4e}$ for not too small $k$ outside that Hamming ball. The likelihood $p(\mc{D}|\pi)$ will favor permutations giving 
best fit to the observed data, so that the posterior mass $\p_{\pi|\mc{D}}(\{\pi: \, \dH(\pi, \textsf{id}) \leq k \}) $ will generally be less than the prior mass. Consequently, it is natural to consider the lower bound $\gamma \sim \log n$ as a starting point when setting this parameter. 
\vskip1ex
\noindent {\bfseries Local shuffling prior}. The next proposition addresses scenarios similar to that
depicted in Figure \ref{fig:motivation_local} when the underlying function is Lipschitz continuous. Specifically, the level of penalty that is needed for the MAP solution to satisfy the bandwidth condition $|\wh{\pi}(i) - i| \leq r$ for all $1 \leq i \leq n$ is provided.
\begin{prop}\label{prop:banded} Suppose that 
$y_i = \mu_{\pi^*(i)} + \sigma_* \eps_i$, $1 \leq i \leq n$, with  $\{ \eps_i \}_{i = 1}^n \overset{\text{i.i.d.}}{\sim}  N(0, 1)$, where $\mu_i = f^*(i/n)$, $1 \leq i \leq n$ for a function $f^*: [0,1] \rightarrow \R$ that is $L$-Lipschitz.    
Let further the matrix $M$ in the prior \eqref{eq:prior} have entries $M_{ij} = I(|i - j| \leq r)$, $1 \leq i,j \leq n$, for
a given bandwidth $r = \max_{1 \leq i \leq n} |\pi^*(i) - i|$. If $\gamma > \frac{2L}{\sigma^*}  \left(\sqrt{\log n} + \sqrt{2} r  \right)$, the resulting MAP estimator $\wh{\pi}$ satisfies $|\wh{\pi}(i) - i| \leq r$, $1 \leq i \leq n$, with probability at least $1 - \exp(-(\sqrt{2} - 1)^2/2) - 2/n$. In particular, under the same event, $\max_{1 \leq i \leq n} |\mu_{\wh{\pi}(i)} - \mu_{\pi^*(i)}| \leq 2L \cdot r / n$. 
\end{prop}
Note that while in theory, the assertion of the above proposition can always be achieved by setting $\gamma = \infty$, established solvers of linear assignment problems require the entries of the cost matrix to be finite. In addition, solver
accuracy can degrade with the magnitude of the entries \cite{Bernhard2021}. 

\section{Experiments}\label{sec:experiments}
In this section, we present the results of experiments conducted with synthetic and real data. At the end of the section, we also show an example demonstrating the use of the data augmentation approach 
discussed at the end of $\S$\ref{subsec:dataproblems} as an alternative to the Monte-Carlo EM scheme. The \texttt{MATLAB} code underlying the results shown herein can be accessed from \url{https://github.com/bruce1edward/exp_prior_shuffled_data}. 
\vskip2ex
\noindent {\bfseries Synthetic data}. We consider data generation according to the following three models:\\[1ex] 
\begin{tabular}{ll}
 Linear Regression (\texttt{LR}):  & $y_{i}|\M{x_{\pi^*(i)}} \sim N(\x^{\T}_{\pi^*(i)}\beta^*, \sigma_*^2)$, $1 \leq i \leq n$,\\[1.3ex]
 Poisson Regression (\texttt{GLM}): & $y_{i}|\M{x_{\pi^*(i)}} \sim \text{Poisson}(\exp(\x^{\T}_{\pi^*(i)}\beta^* + \beta_0^*))$, $1 \leq i \leq n$, \\[1.3ex]
 Multivariate Normal (\texttt{MVN}): & $\M{z}_{i} = (\M{x}_{\pi^*(i)}, \M{y}_{i}) \sim N(\mu_*, \Omega_*^{-1})$, $1 \leq i \leq n$.
 \end{tabular}\\[1.5ex]
The $\{\x_{i}\}^{n}_{i=1}$ and $\{\epsilon_{i}\}^{n}_{i=1}$ are i.i.d random samples from the $N(0, I_{d})$ and $N(0, 1)$ distributions, respectively. The regression parameter $\beta^*$ is sampled uniformly from the sphere of radius $3$, i.e., the set $\{\beta \in \R^d: \, \nnorm{\beta}_2 = 3\}$, and $\beta_0^* \sim N(0,1)$. For \texttt{MVN}, we let $\mu_* = 0$ and $\Omega_*^{-1} = (\tau_* - \rho_*)I_{p+q} + \rho_*\M{1}_{p+q}\M{1}^{\T}_{p+q}$. Finally, $\pi^*$ is a permutation selected uniformly at random from one of the following three constraint sets:
\begin{align}
&\text{(i)}\;k\text{-\textbf{Sparse}:}\, \left\{\pi \in \mc{P}(n) : \sum^{n}_{i=1}\mathbb{I}(\pi(i) \neq i) \leq k \right\}, \;\text{(ii)}\;r\text{-\textbf{Banded}:}\, \Big\{ \pi \in \mc{P}(n): \; \max_{1 \leq i \leq n} |\pi(i) - i| \leq r \Big\}, \notag \\
&\text{(iii)}\;k\text{-\textbf{SparseBlock}:}\, \left\{\pi \in \mc{P}(B) : \sum^{n}_{i=1}\mathbb{I}(\pi(i) \neq i) \leq k \cdot B\right\}, \label{eq:experiments_constraints}
\end{align}
where $\mc{P}(B)$ denotes the set of block-structured permutations corresponding
to $B$ blocks of uniform size $n/B$, i.e., $\{1,\ldots,n/B \}$, $\{n/B + 1, \ldots, 2\cdot n/B\}, \ldots, \{(B-1) (n/B) +1,\ldots,  n \}$. Note that in (i), the number $k$ refers to the number
total of mismatches, whereas in (iii) the number $k$ refers to the number of mismatches 
per block. 
We fix $n = 1,000$, $d = 20$, $\sigma_* = 1$, $\rho_* = 0.8$, $p = q = 5$, $B = 50$. The mismatch rates $k/n$ and $k \cdot B / n$ in \eqref{eq:experiments_constraints}(i) and \eqref{eq:experiments_constraints}(iii), respectively, are varied between $0.2$ and $0.5$ in steps of $0.05$, and the bandwidth $r$ in \eqref{eq:experiments_constraints}(ii) is varied between $3$ and $10$. For each setup and each value of $k$ and $r$, 100 independent replications are performed. 

\vskip1ex
\noindent The following approaches are compared: \par
\noindent (I) {\textsf{naive}}. Standard maximum likelihood estimation as used for parameter estimation in the absence of
mismatches, which corresponds to fixing $\pi = \textsf{id}$ as the identity permutation.
\vskip1ex
\noindent (II) {\textsf{oracle}}. The unknown permutation $\pi^*$ is considered as known, and standard maximum likelihood estimation is used for parameter estimation after fixing $\pi = \pi^*$. 
\vskip1ex
\noindent (III) {\textsf{robust}} [for setting $k$-{\bfseries Sparse} only]. For setup \texttt{LR}, the regression parameter is 
estimated on the \texttt{robustfit} function in \texttt{MATLAB} \cite{MATLAB:2019}. For setup \texttt{GLM}, the regression parameter is estimated based on the robust GLM estimation method \cite{wang2020estimation} that uses observation-specific dummy variables and penalization. For
setup \texttt{MVN}, $\Omega_*^{-1}$ is estimated according to the \texttt{robustcov} function in \texttt{MATLAB} \cite{MATLAB:2019} which implements the minimum covariance determinant estimator in \cite{rousseeuw1999fast}. 
\vskip1ex
\noindent (IV) {\textsf{EM}}, {\textsf{EMH}}, {\textsf{EML}}, {\textsf{EMB}}.  
Algorithm \ref{alg:montecarloEM} using uniform, Hamming, local shuffling, and block-Hamming prior, respectively, where the latter three reflect the constraint sets (i) to (iii) in \eqref{eq:experiments_constraints}. The EM iterations are initialized by setting 
$\pi = \textsf{id}$, and the number of EM iterations is limited to 400. The number of MCMC iterations per EM iteration is set to 8k, half of which are counted towards the ``burn-in period". We note that a modified MH algorithm is used for {\textsf{EML}} (cf.~Appendix \ref{app:MH_local}); for \textsf{EMB}, the MH scheme in Algorithm \ref{alg:MH} is applied block by block. For the {\bfseries Sparse} and {\bfseries SparseBlock} settings, the hyperparameter $\gamma$ is chosen based on Proposition \ref{prop:conc_inequality}, which suggests $\gamma \propto \log(n)$. For the {\bfseries Banded} setting, the choice $\gamma = 1$ was found to yield good performance.  
\vskip1ex
\noindent (V) {\textsf{Lahiri \& Larsen (LL)}}, {\textsf{Chambers (C)}} [for setting $k$\text{-\textbf{SparseBlock}} only]. The approaches described in \cite{Chambers2009} and \cite{Lahiri05} with the choice $Q = \E[\Pi^*] = I_B \otimes Q_{0}$, 
where $Q_{0} = (1 - \alpha_*) I_{n/B} + \alpha_* \M{1}_{n/B}\M{1}_{n/B}^{\T}$,
$\alpha_* = (k \cdot B)/n$. For setup \texttt{MVN}, the LL approach amounts 
to estimation of $\Omega_*$ by the inverse of the modified sample covariance
matrix $\wt{S}$ with blocks
$\wt{S}_{\M{x}\M{x}} = \M{X}^{\T}\M{X} / n$, $\wt{S}_{\M{x}\M{y}} = \M{X}^{\T}Q^{\T}\M{Y} / n$, and $\wt{S}_{\M{y}\M{y}} =\M{Y}^{\T}\M{Y} / n$.

\vskip1ex
\noindent (VI) {\textsf{Averaging}} [for setting $r$-{\bfseries Banded} only]. We compute (componentwise) running averages of the $\{ \x_i \}_{i = 1}^n$ and $\{ \M{y}_i \}_{i = 1}^n$ within sliding windows of size $r$, and estimate the parameters $\beta^*$ and $\Omega_*$ from these local averages in the usual way (i.e., if these were the original and uncontaminated data). \par 
\vskip1ex
\begin{figure}[!htbp]
\begin{center}
\begin{tabular}{ccc}
  \hspace*{-2.5ex} Hamming \textbf{LR} &  Hamming \textbf{GLM} &  Hamming \textbf{MVN}\\
  \hspace*{-2.5ex}  \includegraphics[width = 0.32\textwidth]{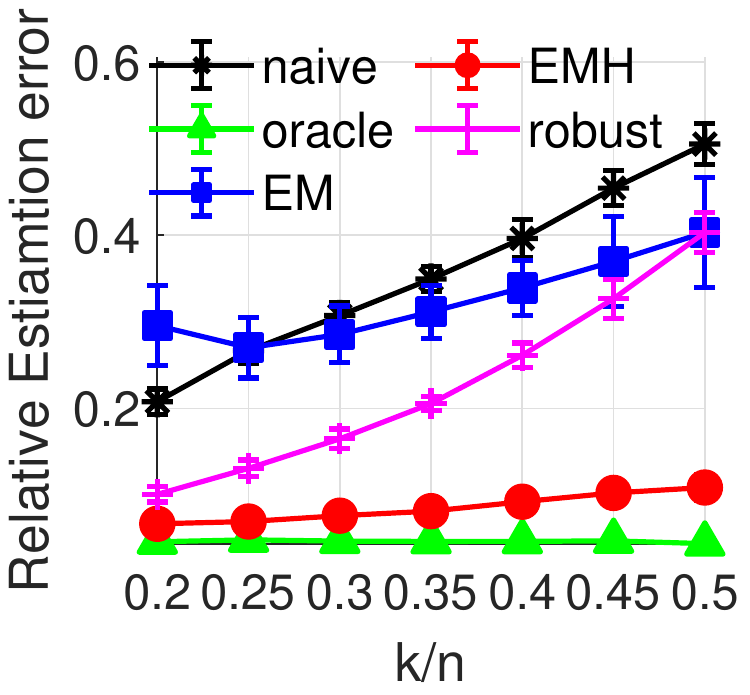} 
     & \includegraphics[width = 0.32\textwidth]{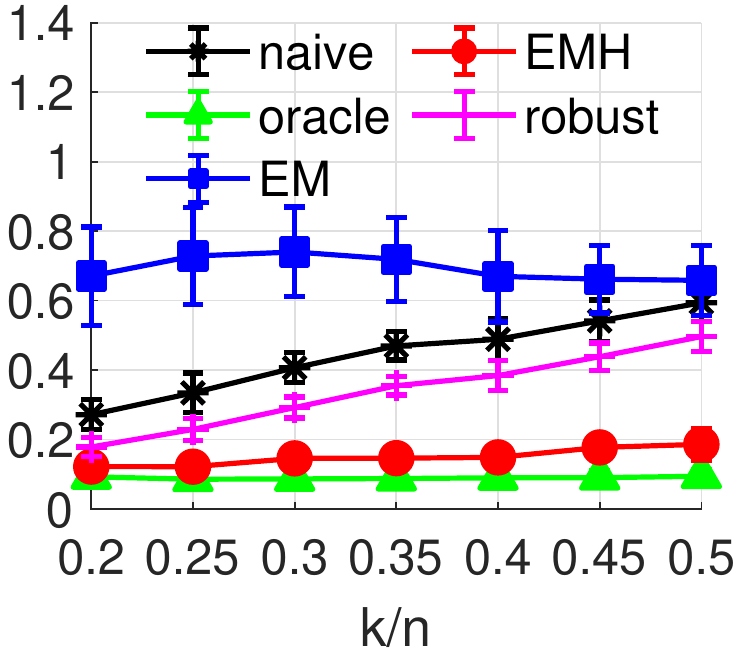}
     & \includegraphics[width = 0.32\textwidth]{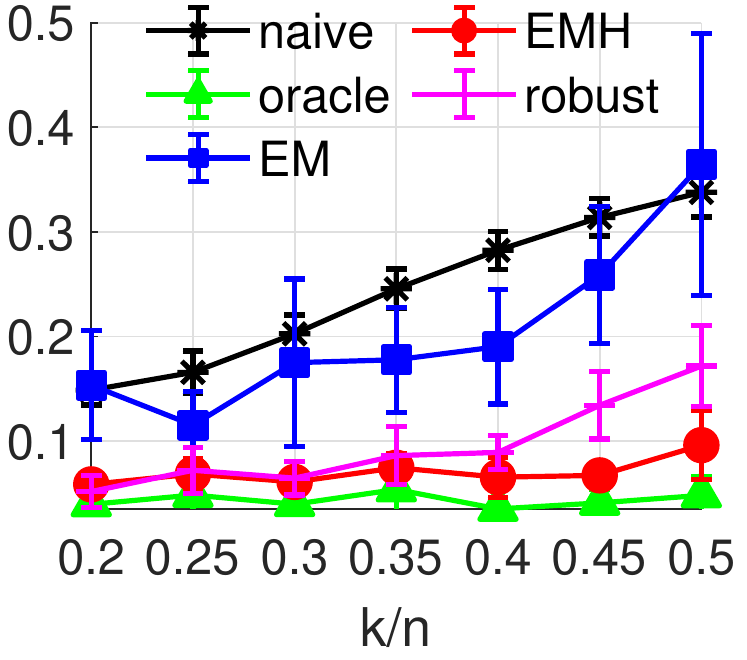}
\end{tabular}
\vskip3ex
\begin{tabular}{ccc}
\hspace*{-2.5ex} SparseBlock \textbf{LR} &  SparseBlock \textbf{GLM} & SparseBlock \textbf{MVN} \\
\hspace*{-2.5ex}  \includegraphics[width = 0.32\textwidth]{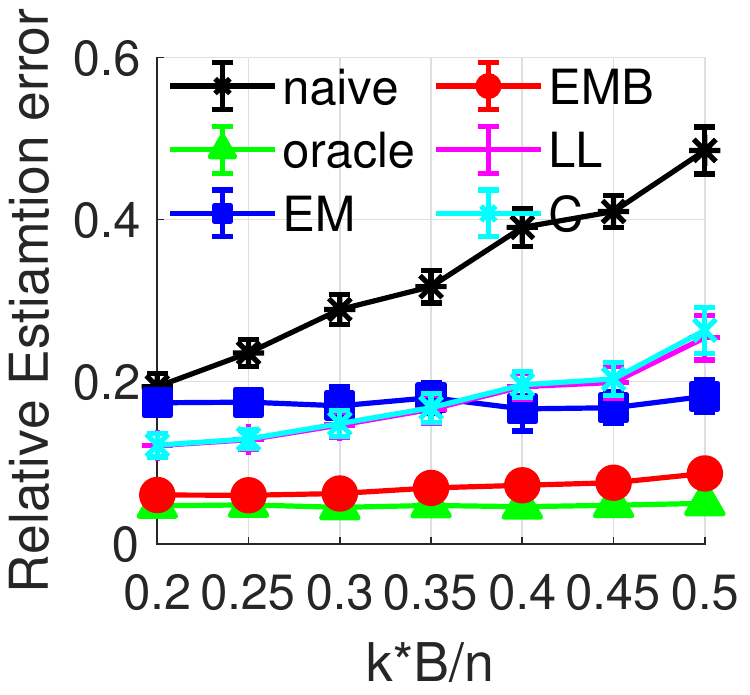} 
    & \includegraphics[width = 0.32\textwidth]{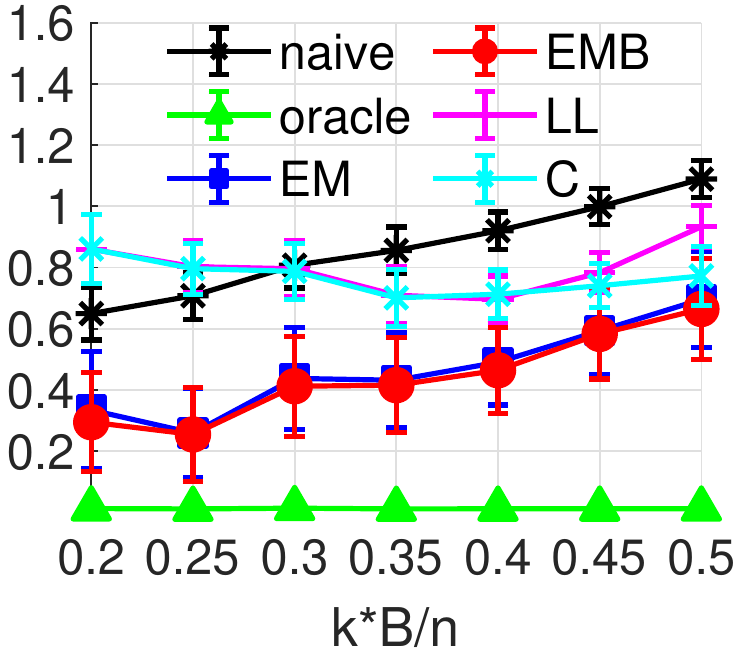}
    & \includegraphics[width = 0.32\textwidth]{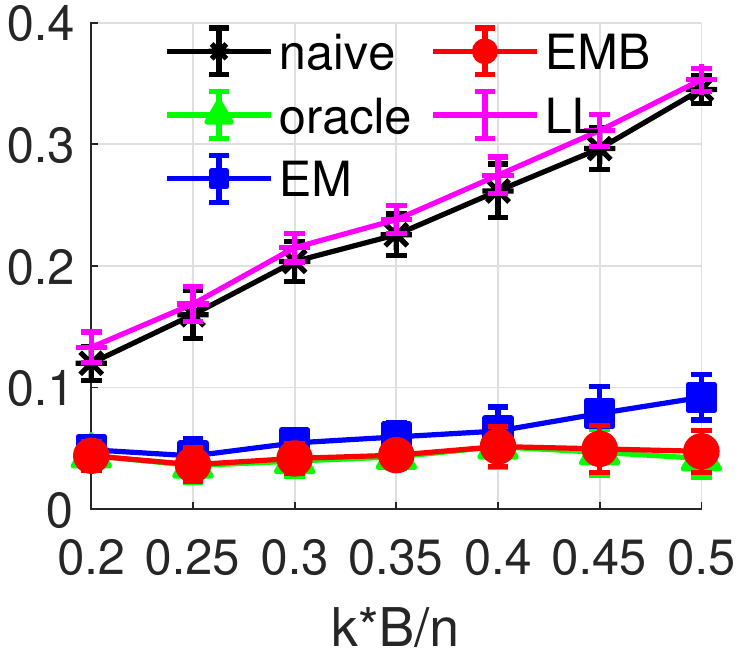}
\end{tabular}
\vskip3ex
\begin{tabular}{ccc}
\hspace*{-2.5ex} Local \textbf{LR} & Local \textbf{GLM} & Local \textbf{MVN}\\
\hspace*{-2.5ex}  \includegraphics[width = 0.32\textwidth]{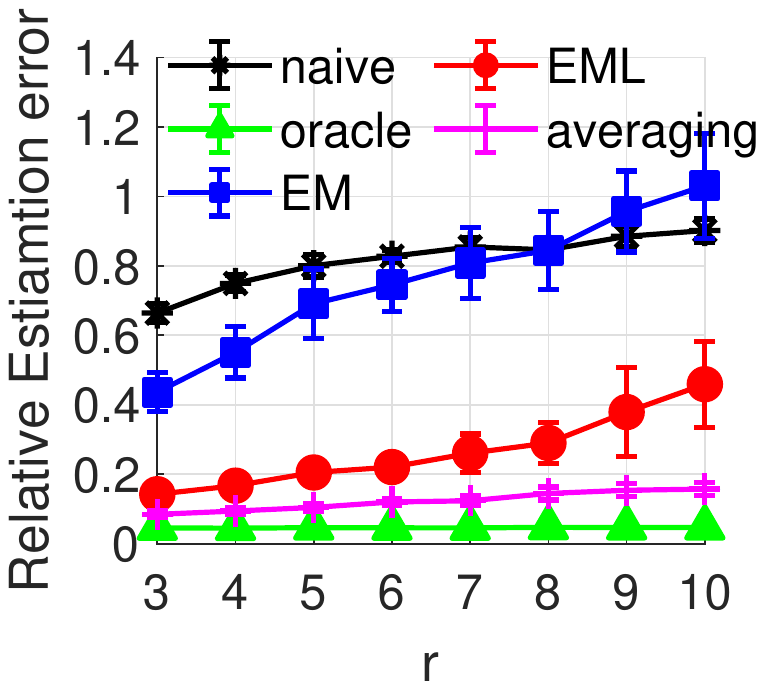}
    & \includegraphics[width = 0.32\textwidth]{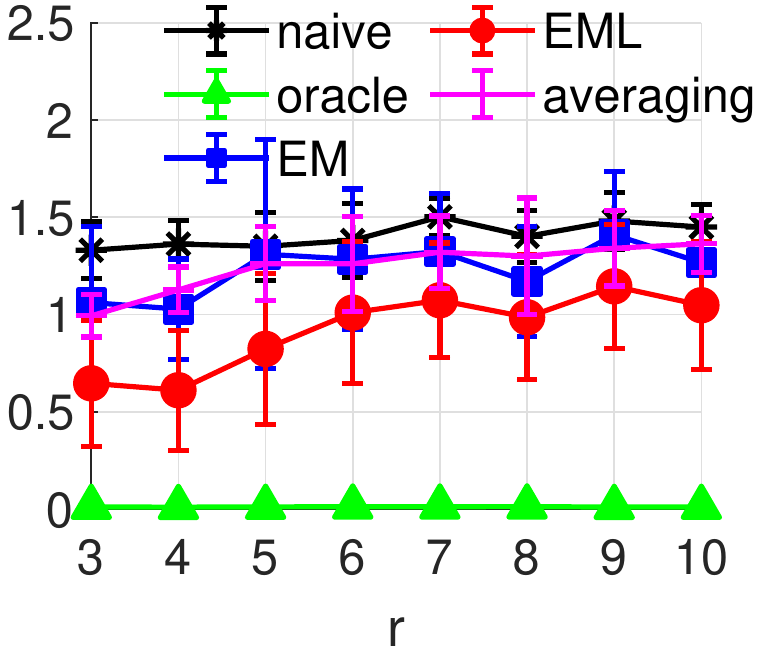}
    & \includegraphics[width = 0.32\textwidth]{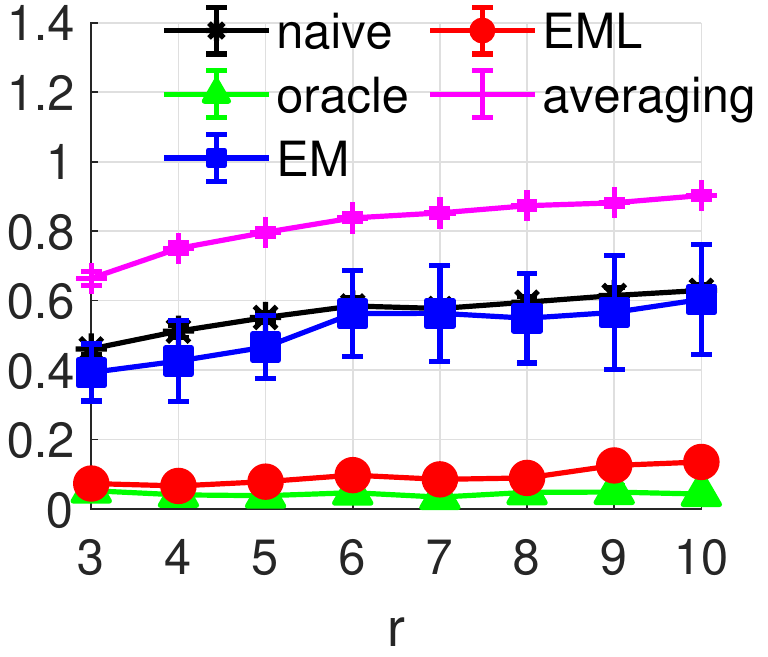}
\end{tabular}
\caption{Results of the synthetic data experiments. The corresponding error bar represent $\pm$3 $\times$ standard error. The figure captions represent the prior and model (boldface) under consideration.}\label{fig:sim}
\end{center}
\end{figure}
For better comparison across different experimental configurations, we visualize the relative estimation error (REE) $\nnorm{\beta^{\text{est}} - \beta^*}_{2}/\nnorm{\beta^*}_{2}$ and $\nnorm{\text{Corr}^{\text{est}} - \text{Corr}^*}_{\textsf{F}}$ ,
where $\beta^{\text{est}}$ and $\text{Corr}^{\text{est}}$ are a placeholder for the aforementioned estimators; $\text{Corr}$ refers to the correlation matrix corresponding to $\Omega_*^{-1}$. Selected results are shown in Figure \ref{fig:sim}, which displays averages of the REE over 100 replications for each model and each permutation. Overall, it can be seen that {\textsf{EMB}}, {\textsf{EMH}}, {\textsf{EML}} achieve significant improvements over their unregularized counterpart and the other baseline methods under consideration. \par 
\begin{table}[hb]
\begin{center}
\caption{Overview of the data set used in the real data analysis. The column ``MCMC steps" lists  the total number of mcmc iteration after the burn-in period (within each block).}\label{table:1}
\vskip.5ex
{\small \begin{tabular}{lllllll} \hline\hline
data(abbreviation) & $n$ & $d/p$ & $q$ & model & setting of $\pi$ & $\text{MCMC steps}$   \\
\hline
\text{Italian survey data(ISD)}\;\cite{SlawskiDiaoBenDavid2019} & 2011 & 2 &  & \texttt{LR} & $k$\text{-\textbf{Sparse}} & 2k \\
\text{El Nino Data(END)}\;\cite{SlawskiDiaoBenDavid2019} & 93935 & 5 &  & \texttt{LR} & $k$\text{-\textbf{SparseBlock}} & 1.5k \\
\text{CPS wage data(CPS)}\;\cite{SlawskiDiaoBenDavid2019} & 534 & 11 &  & \texttt{LR} & $k$\text{-\textbf{Sparse}} & 2k\\
\text{Bike sharing data(BSD)}\;\cite{wang2020estimation} & 731 & 16 &  & \texttt{GLM} & $k$\text{-\textbf{SparseBlock}} & 1.5k \\
\text{Flight Ticket Prices(FTP)}\;\cite{SlawskiBenDavidLi2019} & 335 & 30 & 6 & \texttt{MVN} & $k$\text{-\textbf{Sparse}} & 2k \\
\text{Supply Chain Management(SCM)}\;\cite{SlawskiBenDavidLi2019} & 8966 & 35 & 16 & \texttt{MVN} & $k$\text{-\textbf{Sparse}} & 4k \\
\text{Beijing Air Quality data(BAQD)}\;\cite{SlawskiBenDavidLi2019} & 9762 & 5 & 5 & \texttt{MVN} & $r$\text{-\textbf{Banded}} & 2k\\
\hline\hline
\end{tabular}}
\end{center}
\end{table}
\noindent {\bfseries Real data}. We consider seven benchmark data sets for shuffled data problems as tabulated in Table \ref{table:1}. The data sets are preprocessed versions of their original counterparts. The details of data processing can be found in the corresponding reference provided in Table \ref{table:1}. 
Even though the data sets themselves are real, the permutations that scramble the given 
matching pairs $\{ (\M{x}_i, \M{y}_i) \}_{i = 1}^n$ are synthetic. Specifically, for each data set, we consider 100 independent random permutations depending on the underlying setting. We consider the same list of competitors and associated configurations as for the synthetic data experiments. Asterisked ground truth parameters here refer
to oracle estimates based on knowledge of $\pi^*$, and relative estimation error
(REE) is defined accordingly in terms of those (pseudo)-ground truth parameters.\par

\begin{figure}[!htbp]
\begin{center}
\begin{tabular}{ccc}\label{fig:real_data}
    Hamming ISD &  SparseBlock END & Hamming FTP \\
    \hspace*{-1.5ex}\includegraphics[width = 0.32\textwidth]{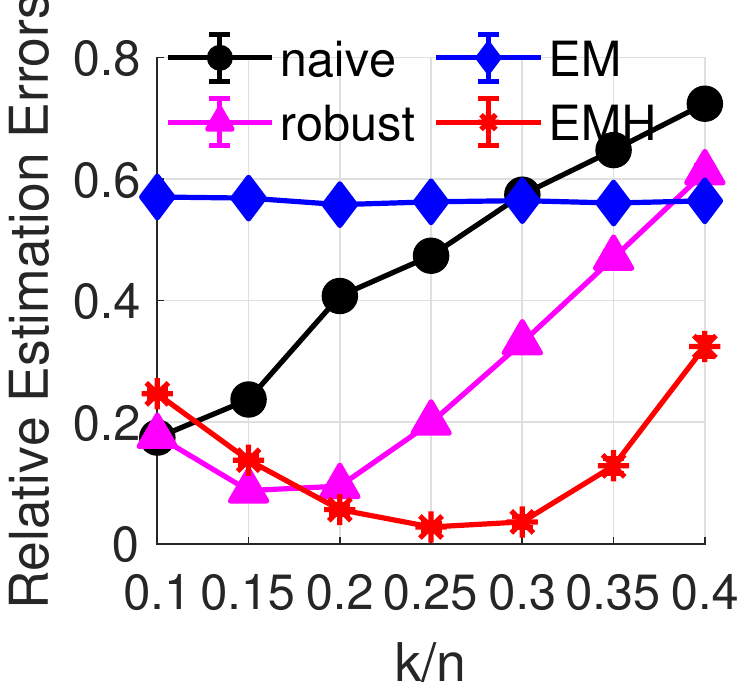}
    & \hspace*{-1.5ex} \includegraphics[width = 0.32\textwidth]{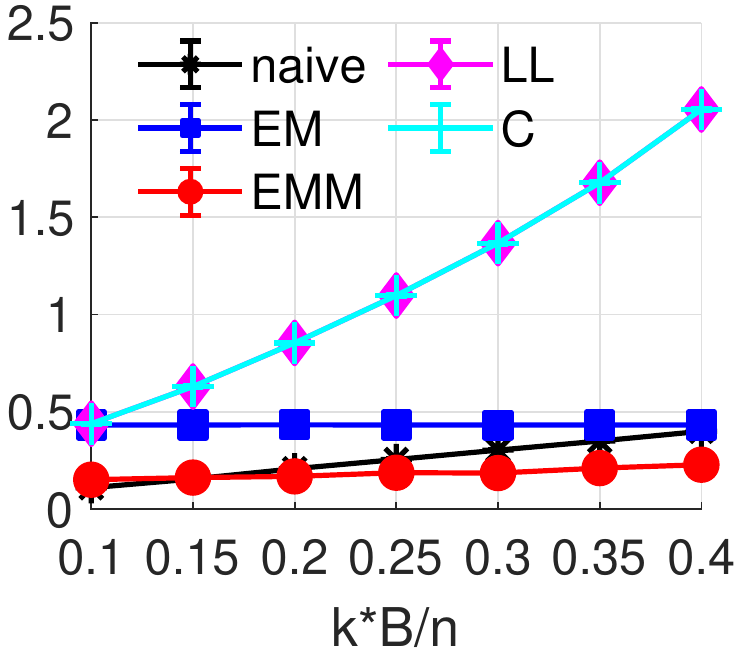} & \hspace*{-1.5ex} \includegraphics[width = 0.32\textwidth]{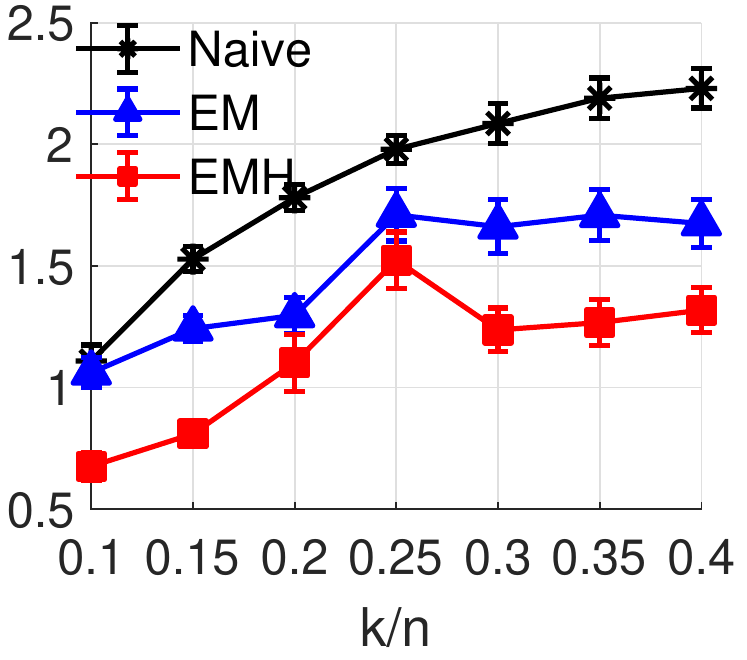} \\[2ex]
    Hamming CPS & SparseBlock BSD & Hamming SCM\\
    \hspace*{-1.5ex}\includegraphics[width = 0.32\textwidth]{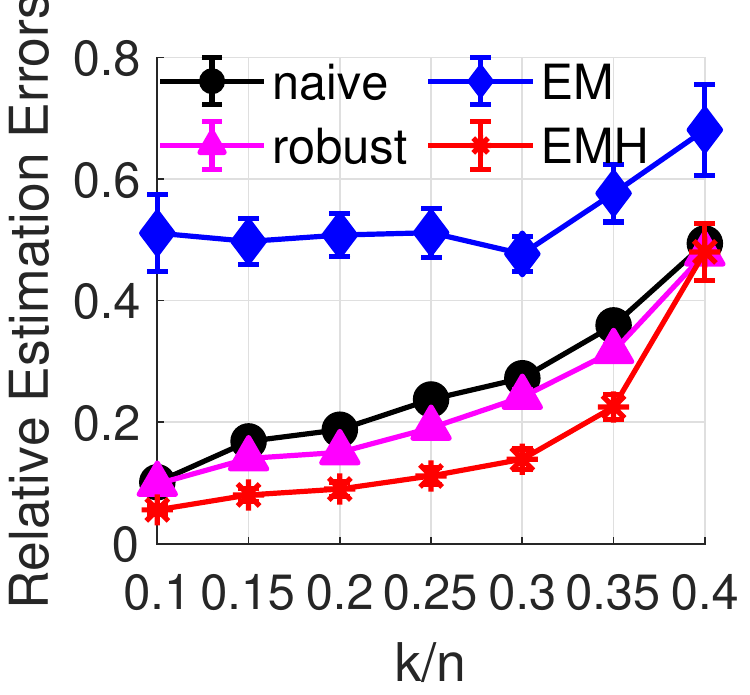} & \hspace*{-1.5ex}\includegraphics[width = 0.32\textwidth]{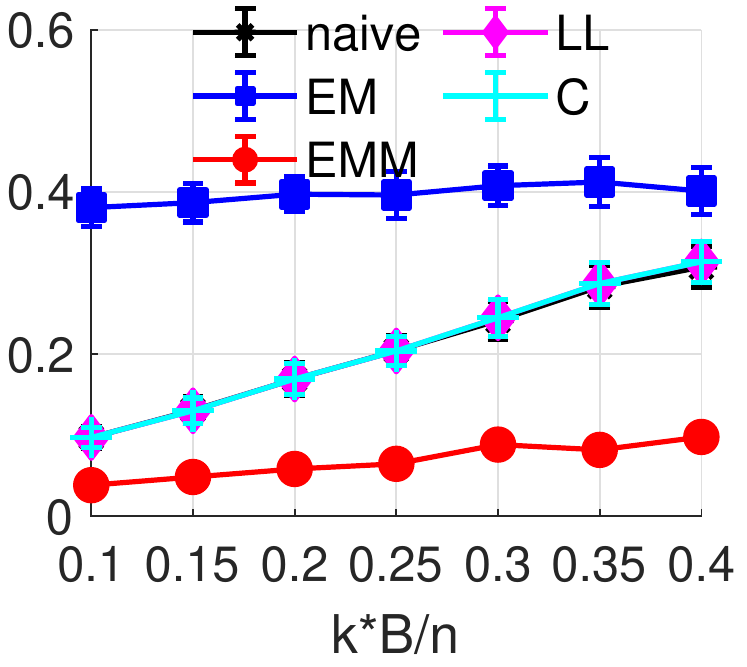} & \hspace*{-1.5ex}\includegraphics[width = 0.32\textwidth]{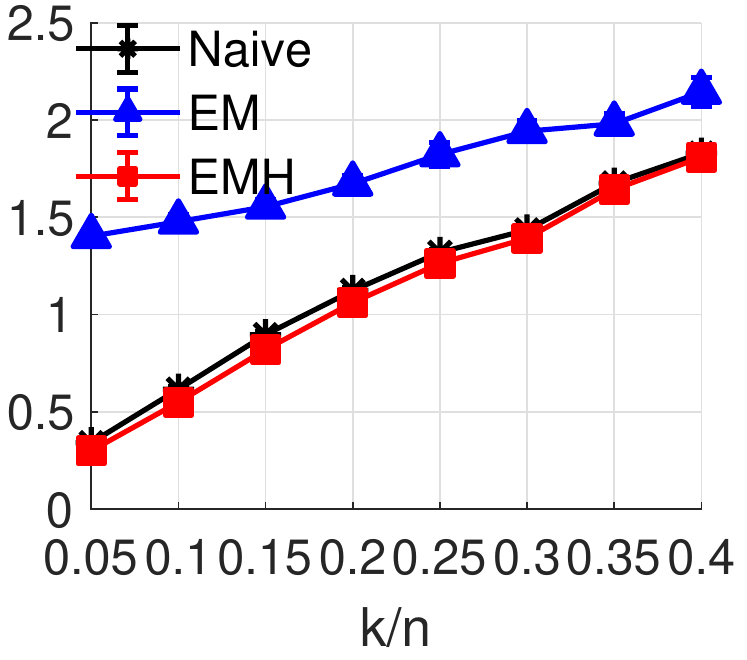}
\end{tabular}
\caption{Results of the real data experiments. The corresponding error bars represent $\pm$3 $\times$ standard error. The figure captions represent the prior and data set (cf.~Table \ref{table:1}) under consideration.}
\end{center}
\end{figure}

\noindent {\bfseries Hamming and Block prior}.  As can be seen from Figure \ref{fig:real_data}, the proposed approach consistently improves over naive least squares once the fraction of mismatches exceeds 0.2, and yields significant improvements as that fraction increases. The regularized EM approach with specific prior (i.e. {\textsf{EMB}}, {\textsf{EMH}}, {\textsf{EML}}) noticeable reduces error induced by shuffling. \par 
\vskip1ex
\noindent {\bfseries Local shuffling prior.}  As shown in the Table \ref{table: 2}, the EM approach with local shuffling prior achieves significant error reductions compared to the naive approach and the EM approach without regularization.  
\begin{table}[h]
\caption{Results of the real data experiment (Beijing Air Quality Data) with local shuffling permutation. Each number in the table is the average REE over 100 replications.} \label{table: 2}
\begin{center}
\begin{tabular}{cccc}
\textbf{Methods} & \textbf{naive} & \textbf{EM} & \textbf{EML} \\
\hline \\[-2ex]
$\nnorm{\text{Corr}^{\text{est}} - \text{Corr}^*}_{F}$ & 0.76 & 1.97 & 0.34 \\
standard error & 0.0012 & 0.0111 & 0.0010
\end{tabular}
\end{center}
\end{table}

\subsection*{Data Augmentation}
In this paragraph we present a brief illustration of the proposed approach when used
in conjunction with data augmentation, i.e., both the parameter and the permutation
are sampled in an alternating fashion (cf.~discussion at the end of $\S$\ref{subsec:dataproblems}). For this purpose, we consider the Italian household survey discussed in \cite{Tancredi15}, see also Table \ref{table:1}. This data set involves a simple linear 
regression problem in which the household income (in 1k Euros) in 2010 is  
is regressed on the same quantity in 2008, including an intercept term. 

The process of file linkage subject to mismatch error involving the income 
data from the two years under consideration is simulated by generating 
a permutation $\pi^*$ uniformly at random from the Hamming ball of 
radius $k$ around the identity permutation, where $k/n = 0.4$. 
 
Following the paradigm of data augmentation in \cite{TannerWong1987} in which
$\pi^*$ is considered as missing data yields the following scheme that
alternates between sampling of a permutation $\pi$ and sampling of regression parameters
$\beta = (\beta_0, \beta_1)$ and $\sigma^2$ given responses $\M{Y} = (y_i)_{i = 1}^n$ (income from 2010) and design matrix $\M{X} = [\bm{1}_n \; (x_i)_{i = 1}^n]$ (intercept and income from 2008).
\begin{alignat*}{2}
    &\text{(I) Augmentation Step:} \;\;\, && \text{Sample} \; \pi^{(j)} \text{ from } p(\pi|\M{Y},\M{X}, \beta^{(k-1)}, {\sigma^2}^{(k-1)}), \; \, j = 1, \ldots, m,\\
    &\text{(II) Posterior Step:}\;\;\, && \text{(a) Sample } \beta^{(k)} \; \text{from}  \; \, \frac{1}{m}\sum_{j=1}^{m}p(\beta|{\sigma^2}^{(k-1)}, \pi^{(j)}, \M{Y},\M{X}), \\
    &   && \text{(b) Sample} \; {\sigma^2}^{(k)} \text{ from } \;\,  \frac{1}{m}\sum_{j=1}^{m}p(\sigma^2|\beta^{(k)}, \pi^{(j)}, \M{Y},\M{X}),
\end{alignat*} 
where $m$ denotes the number of samples in the augmentation step, and $k$ denotes the iteration counter for the parameters $(\beta, \sigma^2)$. 

Sampling in step (I) is implemented according to the MH procedure shown in Algorithm \ref{alg:MH}. Furthermore, under the usual non-informative prior distribution for $(\beta, \sigma^2)$, i.e., $p(\beta, \sigma^2) \propto \sigma^{-2}$, the full conditonal distributions appearing in step (II) are given by 
\begin{align*}
&\beta|\sigma^2, \Pi, \M{Y}, \M{X} \sim N(\wt{\beta}, \sigma^2 (\M{X}^{\T} \M{X})^{-1}), \quad \sigma^2|\beta, \Pi, \M{Y}, \M{X} \sim \text{Inv-}\chi^2(n - d, s^2), \\
&\wt{\beta} \coloneq (\M{X}^{\T} \M{X})^{-1} \M{X}^{\T} \Pi^{\T} \M{Y}, \quad  s^2 \coloneq \frac{1}{n-d} \nnorm{\M{Y} - \Pi\M{X}\wt{\beta}}_2^2,
\end{align*}
where $\text{Inv-}\chi^2(\nu, a^2)$ refers to the scaled inverse $\chi^2$-distribution
with scale parameter $a > 0$ and $\nu$ degrees of freedom (cf.~$\S$14 in \cite{gelman2013bayesian} for more details on Bayesian inference for linear regression models).


For this illustration, we use $m = 100$, where each sequence $\{ \pi^{(j)} \}$ is 
generated by uniform thinning of Markov chains of length $4,000$ generated 
by Algorithm \ref{alg:MH}. The number of samples $(\beta^{(k)}, {\sigma^2}^{(k)})$
obtained via the above scheme is taken as 1,000. The sampling procedure is initialized
from step (II) with the identity permutation. We compare both the unregularized
case with the uniform prior for $\pi$ as well as the regularized case with the 
Hamming prior \eqref{eq:Hammingprior} ($\gamma = \log n$ in view of Proposition \ref{prop:conc_inequality}). 

\begin{figure}[!htbp]
\begin{center}
\begin{tabular}{ccc}
     \includegraphics[width = 0.3\textwidth]{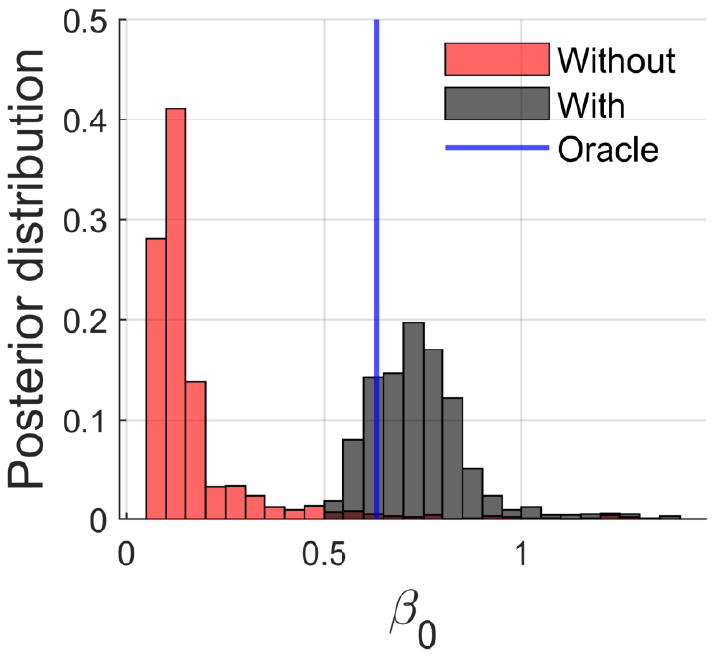}
    & \includegraphics[width = 0.3\textwidth]{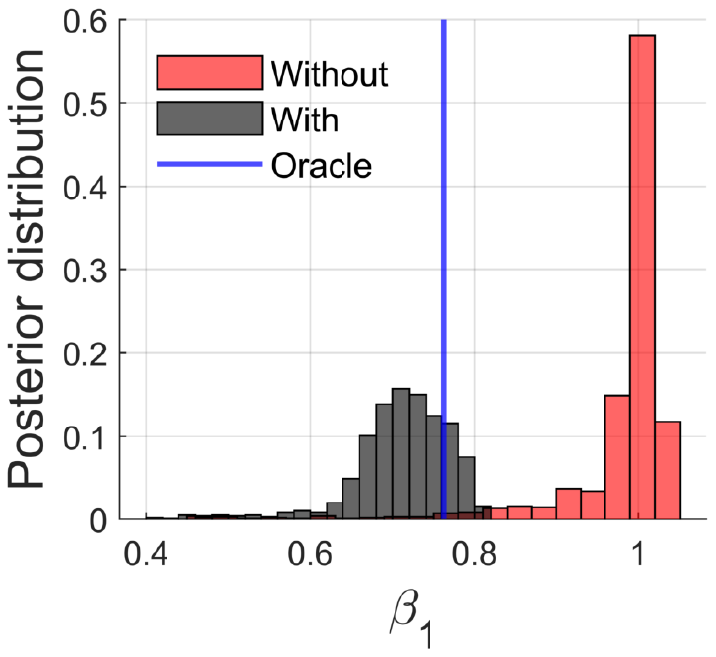} 
    & \includegraphics[width = 0.3\textwidth]{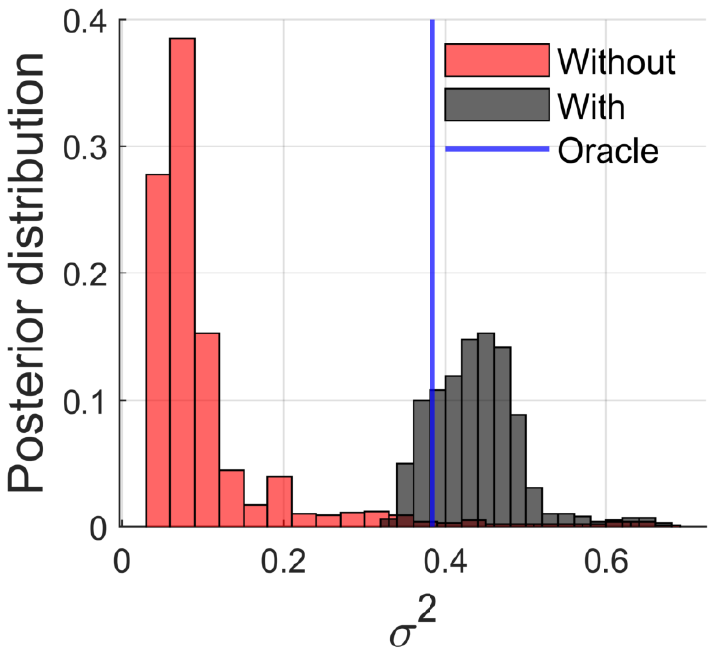}
\end{tabular}
\vspace*{-2ex}
\caption{Comparison between the posterior distributions for the parameters
$(\beta_0, \beta_1, \sigma^2)$ for the Italian household survey data (i) with regularization based on the proposed Hamming prior for $\pi$ ($\gamma = \log n$, grey histogram) and without regularization, i.e., uniform prior for $\pi$ (red histogram). ``Oracle" refers to the least squares estimator in the absence of mismatch error.}\label{fig: data_augmentation}
\end{center}
\end{figure}


Figure \ref{fig: data_augmentation} confirms that the proposed approach achieves
visible improvements over the unregularized approach which suffers from serious
amplification bias affecting the slope parameter $\beta_1$ and serious underestimation
of the error variance,  as predicted by the brief analysis accompanying
the first introductory example in $\S$\ref{subsec:mot_example}. 



\section{Conclusion}\label{sec:conclusion}
In this paper, we have proposed a framework for regularized estimation in shuffled data problems by means
of an exponential family prior on the permutation group. As elaborated above, the exponential family 
form is convenient for computational purposes yet sufficiently rich to incorporate various common
forms of prior knowledge. In particular, the prior is not tailored to specific data analysis problems,
but can be applied generically. The results in this paper confirm the importance of regularization
in shuffled data problems given the inherent danger of overfitting already in the presence of little
noise. While the approach covers a variety of constraints that can be imposed on the underlying permutation, it is 
certainly not exhaustive: not all sorts of prior knowledge can be captured via the prior proposed herein. For 
example, suppose we have information on the cycles of the permutation (numbers and/or lengths). Such information
cannot be expressed in terms of index pairs, and hence requires a different paradigm. Several authors 
\cite{Kondor2007, Huang2009} have considered the Fourier analysis on the permutation group \cite{Diaconis1988} to facilitate learning problems 
involving permutations, and it is an interesting direction of future research to study how 
that approach can be leveraged for the type of shuffled data problems considered in the present paper. Moreover, we herein have focused on the situation where exactly two files are merged. It is of interest to consider more complex situations arising from linkage of several files and multiple 
permutations as well as the modeling of potential dependencies among those: for examples, the permutations might be completely unrelated, or identical \cite{SlawskiBenDavidLi2019, ZhangSlawskiLi2019}, or involve intermediate situations with varying degrees of overlap. 


\bibliographystyle{IEEEtran}
\bibliography{refs.bib}

\appendix

\section{Proof of \eqref{eq:convergence_overfitting}}\label{app:overfitting_ml} 
Let $\mu_i = x_i \beta^*$, $1 \leq i \leq n$, and let $P_{\mu}^n$ and $P_y^n$ be the probability measures with mass
$1/n$ at the $\{ \mu \}_{i = 1}^n$ and $\{ y_i \}_{i = 1}^n$, respectively. Then the squared Wasserstein-2 distance  
$\textsf{W}_2^2$ between $P_{\mu}^n$ and $P_{y}^n$ is given by \cite[cf.][$\S$2.3]{COT2019}
\begin{align}
\textsf{W}_2^2(P_{\mu}^n, P_y^n) = \min_{\pi \in \mc{P}(n)} \frac{1}{n} \su \{y_i - \mu_{\pi(i)} \}^2 &=
\frac{1}{n} \su y_i^2 + \frac{1}{n} \su \mu_i^2 - \frac{2}{n} \left\{ \max_{\pi \in \mc{P}(n)} \su x_i y_i \right\} \beta^* \notag \\
&= \frac{1}{n} \su y_i^2 + \frac{1}{n} \su \mu_i^2 - \frac{2}{n} \su x_{(i)} y_{(i)} \beta^* \label{eq:Wasserstein}
\end{align}
We have that $\textsf{W}_2^2(P_{\mu}^n, P_y^n) \rightarrow \textsf{W}_2^2(P_{\mu}, P_y) = (\sqrt{(\beta^*)^2 + \sigma_*^2} - \beta^*)^2$ in probability
as $n \rightarrow \infty$, where $P_{\mu}$ and $P_y$ denote the Gaussian measures $N(0, (\beta^*)^2)$ and $N(0, (\beta^*)^2 + \sigma_*^2)$, respectively \cite[][Remark 2.31]{COT2019}. At the same time, $\frac{1}{n} \su y_i^2 \rightarrow (\beta^*)^2 + \sigma_*^2$ and $\frac{1}{n} \su \mu_i^2 \rightarrow (\beta^*)^2$ in probability as $n \rightarrow \infty$. Substitution into \eqref{eq:Wasserstein} and invoking Slutsky's theorem, we have that
\begin{equation*}
\frac{1}{n} \su x_{(i)} y_{(i)} \rightarrow \sqrt{(\beta^*)^2 + \sigma_*^2}. 
\end{equation*}
in probability as $n \rightarrow \infty$. The first result in \eqref{eq:convergence_overfitting} then follows immediately from Slutsky's Theorem and the fact that $n^{-1} \su x_i^2 \rightarrow 1$, and observe that the third result in \eqref{eq:convergence_overfitting} is obtained as a direct consequence with the same reasoning. The result $\wh{\sigma}_{\text{ML}}^2 \rightarrow 0$ is obtained by expanding the square 
\begin{equation*}
 \frac{1}{n} \su y_i^2  - \frac{2}{n} \su y_i x_i \wh{\beta}_{\text{ML}}   + \frac{1}{n} \su x_i^2  (\wh{\beta}_{\text{ML}})^2,
\end{equation*}
and analyzing each of the terms accordingly. 

\section{Proof of Theorem \ref{theo:hamming}} 
In light of relation \eqref{eq:proof_conjugacy}, straightforward manipulations and omission of terms not depending on $\Pi$ show that the MAP estimator $\wh{\Pi}$ is the minimizer of the optimization problem
\begin{equation}\label{eq:MAP_representation_linearreg}
\min_{\Pi \in \mc{P}(n)} \left\{ -\scp{\M{Y}}{\Pi \bm{\mu}} +\sigma_*^2 \gamma \dH(\Pi, I_n) \right \}. 
\end{equation}
Since $\wh{\Pi}$ minimizes \eqref{eq:MAP_representation_linearreg}, the following basic inequality holds true:
\begin{equation}\label{eq:basic_inequality_linear}
-\nscp{\M{Y}}{\wh{\Pi} \bm{\mu}} + \sigma_*^2 \gamma \dH(\wh{\Pi}, I_n) \leq -\nscp{\M{Y}}{\Pi^* \bm{\mu}} + \sigma_*^2 \gamma \dH(\Pi^*, I_n)
\end{equation}
Decomposing $\M{Y} = \bm{\mu} + \bm{\xi}$ with $\bm{\xi} = \sigma_* \bm{\eps}$ and re-arranging terms in the above inequality yields that 
\begin{equation*}
\nscp{\Pi^* \bm{\mu}}{(\Pi^* - \wh{\Pi} ) \bm{\mu}} - \nscp{\bm{\xi}}{(\wh{\Pi} - \Pi^*) \bm{\mu}} + \sigma_*^2 \gamma \dH(\wh{\Pi}, I_n) \leq  \sigma_*^2 \gamma k,
\end{equation*}    
where we have substituted $\dH(\Pi^*, I_n) = k$. By the Cauchy-Schwarz inequality, $\nscp{\Pi^* \bm{\mu}}{\wh{\Pi} \bm{\mu}} \leq \nnorm{\Pi^* \bm{\mu}}_2^2$, which implies that the first term in the previous inequality is non-negative. This in turn yields
that 
\begin{equation}\label{eq:basic_inequality_reduced}
- \nscp{\bm{\xi}}{(\wh{\Pi} - \Pi^*) \bm{\mu}} + \sigma_*^2 \gamma \dH(\wh{\Pi}, I_n) \leq \sigma_*^2 \gamma k. 
\end{equation}
In the sequel, we will derive a probabilistic lower bound on the first term on the left hand side. 
    
For any integer $1 \leq s \leq n$, consider the event $\mc{E}_s = \{ \dH(\wh{\Pi}, I_n) \leq s \}$, and
let $\overline{v} = \frac{(\wh{\Pi} - \Pi^*) \bm{\mu}}{2 \nnorm{\beta^*}_2}$. We have that 
\begin{equation*}
\nnorm{\overline{v}}_2 = \sup_{\nnorm{u}_2 \leq 1} \scp{u}{\frac{(\wh{\Pi} - \Pi^*) \bm{\mu}}{2 \nnorm{\beta^*}_2}} 
= \sup_{\nnorm{u}_2 \leq 1} \scp{\frac{(\wh{\Pi} - \Pi^*) u}{2}}{\frac{\bm{\mu}}{\nnorm{\beta^*}_2}}
\end{equation*}
Observe that conditional on $\mc{E}_s$, for any vector $u \in \R^n$, $(\wh{\Pi} - \Pi^*) u$ can have at most $m_s = s + k$ non-zero entries. Moreover,  $\nnorm{(\wh{\Pi} - \Pi^*) u}_2 \leq 2 \nnorm{u}_2$. Finally, note that in light of the setting \eqref{eq:gaussianlinear} under consideration, $\bm{\mu} / \nnorm{\beta^*}_2 \sim N(0, I_n)$. It follows that for any $t > 0$
\begin{equation}\label{eq:sup_gauss_1}
\p(\nnorm{\overline{v}}_2 \geq t | \mc{E}_s) \leq \p \left(\sup_{u \in \mc{B}_0(m_s)} \nscp{u}{g} > t \right), \quad g \sim N(0, I_n). 
\end{equation}
where for any integer $1 \leq \ell \leq n$,  $\mc{B}_0(\ell)$ here denotes the set of all unit vectors having at most $\ell$ non-zero entries. Denote by $w(\mc{B}(\ell)) = \E_{g \sim N(0, I_n)}[\sup_{u \in \mc{B}(\ell)} \nscp{u}{g}]$ the corresponding Gaussian width \cite[][$\S$7.5]{Vershynin2018}. Choosing $t = w(\mc{B}(m_s)) + c_1 \sqrt{2 \log n}$ in \eqref{eq:sup_gauss_1} for $c_1 \geq 1$, standard tail bounds for the suprema of Gaussian processes \cite[][Theorem 5.8]{Boucheron2013} yield
\begin{equation*}
\p(\nnorm{\overline{v}}_2 \geq w(\mc{B}_0(m_s)) + c_1 \sqrt{2 \log n} \; | \mc{E}_s) \leq n^{-c_1^2}.  
\end{equation*}
 Using that $w(\mc{B}_0(m_s)) \leq 4 \sqrt{m_s \log(en/m_s)}$ \cite[e.g.,][Lemma 2.3]{PlanVershynin2013a} and the fact that $s \log(en/s) \geq \log n$ for any
$n \geq s \geq 1$, we have with $c_1 = \sqrt{2}$
\begin{equation}\label{eq:control_GP}
  \p(\nnorm{\overline{v}}_2 \geq 6 \sqrt{m_s \log(en/m_s)} \; | \mc{E}_s) \leq 1/n^2. 
\end{equation}
Combining this with the definition of $\overline{v}$ yields that 
\begin{equation*}
  \p(\nnorm{(\wh{\Pi} - \Pi^*) \bm{\mu}}_2 \geq 12 \nnorm{\beta^*}_2 \sqrt{m_s \log(en/m_s)} \; | \mc{E}_s) \leq 1/n^2.
\end{equation*}
Let now $\tau_s = 12 \nnorm{\beta^*}_2 \sqrt{m_s \log(en/m_s)}$ and $\mc{F}_s = \{ \nnorm{(\wh{\Pi} - \Pi^*) \bm{\mu}}_2 \leq \tau_s \}$, and note that
\begin{align*}
  \p(\nscp{\bm{\xi}}{(\wh{\Pi} - \Pi^*) \bm{\mu}} > t   | \mc{F}_s \cap \mc{E}_s) 
  \leq \p\left(\sup_{v \in \mc{B}_0(m_s)} \nscp{g}{v}  \sigma_* \tau_s > t \right), \quad g \sim N(0, I_n).    
\end{align*}
Using the same argument as above, we choose $t = \sigma_* \tau_s \{ w(\mc{B}_0(m_s)) + c_2 \sqrt{2 \log n} \}$ with $c_2 = \sqrt{2}$. Putting together the pieces as above, we obtain
\begin{equation*}
\p(\nscp{\bm{\xi}}{(\wh{\Pi} - \Pi^*) \bm{\mu}} > \underbrace{12 \cdot 6}_{=72} \cdot \sigma_* \nnorm{\beta^*}_2 m_s \log(e n / m_s)  | \mc{F}_s \cap \mc{E}_s) \leq 1/n^2. 
\end{equation*}
Now let $\gamma_0 = 72 \sqrt{\textsf{SNR}} \log(en/k) \geq 72 \sqrt{\textsf{SNR}} \log(e n / m_s)$ and define the event
\begin{equation*}
\mc{G}_s = \left \{ \frac{\left| \nscp{\bm{\xi}}{(\wh{\Pi} - \Pi^*) \bm{\mu}} \right|}{\sigma_*^2}  \leq \gamma_0  m_s \right \} 
\end{equation*}
Observe that conditional on $\mc{E}_s \cap \mc{G}_s$, the earlier inequality \eqref{eq:basic_inequality_reduced} implies that (recalling that $m_s = s + k$)
\begin{equation*}
-\gamma_0 \sigma_*^2 (s + k)  + \sigma_*^2 \gamma s \leq -\nscp{\bm{\xi}}{(\wh{\Pi} - \Pi^*) \bm{\mu}} + \sigma_*^2 \gamma \dH(\wh{\Pi}, I_n) \leq \sigma_*^2 \gamma k. 
\end{equation*}
Combination of the left and right hand sides and re-arranging terms implies the inequality 
\begin{equation*}
\gamma_0 \geq \gamma \frac{s-k}{s + k}
\end{equation*}
Now for any $s \geq 2k$, the right hand side is lower bounded by $(1/3)\gamma$. This in turn yields a contradiction
whenever $\gamma$ is chosen such that $\gamma > 3 \gamma_0$. In order to conclude that $\dH(\wh{\Pi}, I_n) \leq 2k$ with
the stated probability in that case, it remains to provide a corresponding lower bound on the probability of the event
$\bigcup_{s = 1}^n (\mc{E}_s \cap \mc{G}_s)$, i.e., at least one of the events $\{\mc{E}_s \cap \mc{G}_s \}_{s = 1}^n$ occurs. Since the events inside the union are disjoint, we obtain that 
\begin{align*}
\p \left(\bigcup_{s = 1}^n (\mc{E}_s \cap \mc{G}_s) \right) = \sum_{s = 1}^n \p(\mc{E}_s \cap \mc{G}_s) \geq 
\sum_{s = 1}^n \p(\mc{E}_s \cap \mc{G}_s \cap \mc{F}_s)
&= \sum_{s = 1}^n \underbrace{\p(\mc{G}_s | \mc{E}_s \cap \mc{F}_s)}_{\geq 1 - 1/n^2} \underbrace{\p(\mc{F}_s | \mc{E}_s)}_{\geq 1-1/n^2} \p(\mc{E}_s) \\
&\geq \sum_{s = 1}^n (1 - 2/n^2) \p(\mc{E}_s) \geq 1-2/n. 
\end{align*}
In order to prove the second part of Theorem \ref{theo:hamming}, we first invoke the following basic inequality equivalent to \eqref{eq:basic_inequality_linear}
\begin{equation*}
  \nnorm{\wh{\Pi} \bm{\mu} - \M{Y}}_2^2 + 2 \sigma_*^2 \gamma \dH(\wh{\Pi}, I_n) \leq \nnorm{\Pi^* \bm{\mu} - \M{Y}}_2^2 + 2 \sigma_*^2 \gamma \dH(\Pi^*, I_n).
 \end{equation*}
Expanding the squares and re-arranging yields conditional on $\bigcup_{s = 1}^n (\mc{E}_s \cap \mc{G}_s)$
\begin{align*}
  \nnorm{\wh{\Pi} \bm{\mu} - \Pi^* \bm{\mu}}_2^2 &\leq 2\nscp{\bm{\xi}}{\wh{\Pi} \bm{\mu} - \Pi^* \bm{\mu}} + 2 \sigma_*^2 \gamma (k - s) \\
                                      &\leq  2 \sup_{u \in \mc{B}_0(3k)} \nscp{\bm{\xi}}{u} \nnorm{\Pi^* \bm{\mu} - \wh{\Pi} \bm{\mu}}_2 + 2 \sigma_*^2 \gamma (k-s),  
\end{align*}
where in the second inequality, we have used that if $\gamma > 3\gamma_0$, conditional on on $\bigcup_{s = 1}^n (\mc{E}_s \cap \mc{G}_s)$, it holds that $\dH(\wh{\Pi}, I_n) \leq 2k$. The latter inequality is of the form 
\begin{equation*}
x^2 - 2bx - c \leq 0, \qquad x \coloneq \nnorm{\wh{\Pi} \bm{\mu} - \Pi^* \bm{\mu}}_2, \;\;\, b \coloneq \sup_{u \in \mc{B}_0(3k)} \nscp{\bm{\xi}}{u}, \quad c = 2 \sigma_*^2 \gamma (k-s).  
\end{equation*}
After elementary manipulations, we obtain the inequality $x \leq \sqrt{b^2 + c} +  b \leq 2b + \sqrt{c}$, which translates to 
\begin{align*}
\nnorm{\wh{\Pi} \bm{\mu} - \Pi^* \bm{\mu}}_2 \leq 2  \sup_{u \in \mc{B}_0(3k)} \nscp{\bm{\xi}}{u} + \sigma_* \sqrt{2 \gamma} 
                                             \leq \sigma_* \left(17 \sqrt{k \log(e n / 3k)} + \sqrt{2 \gamma} \right),
\end{align*}
with probability at least $1 - 2/n - 1/n = 1-3/n$, where the term $\sup_{u \in \mc{B}_0(3k)}$ is controlled similarly to 
\eqref{eq:control_GP}. 

\section{Proof of Proposition \ref{prop:conc_inequality}} 
The probability mass function of \eqref{eq:Hammingprior} is given by \cite{fligner1986distance}: 
\begin{equation}\label{eq:PMF_hamming}
 p(\pi) = \frac{\exp(-\gamma \dH(\pi, \textsf{id}))}{\psi(\gamma)}, \qquad  \psi(\gamma) \coloneq n!\exp(-\gamma n)\sum^{n}_{k=0}\frac{(\exp(\gamma) - 1)^{k}}{k!}. 
\end{equation}
In the sequel, let us write $\{ D(\pi) = d \} $ as a shortcut for the event $\{ \dH(\pi, \textsf{id}) = d \}$. We then
have 
\begin{align*}
\p_{\pi \sim p}(D(\pi) \geq k) = \sum_{d = k}^n \frac{\exp(-\gamma d)}{\psi(\gamma)} \binom{n}{d} !d,
\end{align*}
where $!d$ denotes the number of derangements of $\{1,\ldots,d\}$, i.e., the number of permutations $\tau$ of 
$d$ objects such that $\tau(j) \neq j$ for all $1 \leq j \leq d$. Straightforward manipulations yield
\begin{align}\label{eq:tailprobability}
    \p(D(\pi) \geq k) = \sum^{n}_{d = k} \frac{\exp(-\gamma d)\frac{n!}{d!(n-d)!}!d}{n!\exp(-\gamma n)\sum^{n}_{\ell=0}\frac{(\exp(\gamma) - 1)^{\ell}}{\ell!}} 
     = \sum^{n}_{d = k} \frac{\exp(\gamma(n- d))}{(n-d)!\sum^{n}_{\ell=0}\frac{(\exp(\gamma) - 1)^{\ell}}{\ell!}} \frac{!d}{d!}. 
\end{align}
For $x \geq 0$ and integer $m \geq 1$, define the (upper) incomplete Gamma function and its ``normalized" counterpart by
\begin{equation*}
\Gamma(m,x) = \int^{\infty}_{x}t^{n-1}e^{-t} \, dt, \qquad \wt{\Gamma}(m,x) = \Gamma(m,x)/\Gamma(m),
\end{equation*}
where $\Gamma(m) = \Gamma(m,0) = (m-1)!$ denotes the Gamma function. 
It can be shown that \cite[][$\S$6.5]{AbramowitzStegun}
\begin{equation}\label{eq:poisson_incomplete_gamma}
\sum^{m}_{k = 0} \frac{x^{k}}{k!} = e^{x} \wt{\Gamma}(m+1, x). 
\end{equation} 
%
Further note that for $k \leq d \leq n$, we have that $\frac{!k}{k!} \leq !d/d! \leq !n/n! \leq e^{-1}$. Accordingly, for 
$\frac{!k}{k!} \leq c_0(n,k) \leq !n/n!$, we obtain the following for the right hand side of \eqref{eq:tailprobability}: 
\begin{align*}
    c_0(n,k) \sum^{n}_{d = k} \frac{\exp(\gamma(n- d))}{(n-d)!\sum^{n}_{\ell=0}\frac{(\exp(\gamma) - 1)^{\ell}}{\ell!}} 
    &= c_0(n,k) \sum^{n - k}_{i = 0} \frac{(\exp(\gamma))^i}{i!} \frac{1}{\sum^{n}_{\ell=0}\frac{(\exp(\gamma) - 1)^{\ell}}{\ell!}} \\
    &= c_0(n,k) \frac{e^{\exp(\gamma)} \, \wt{\Gamma}(n - k + 1, \exp(\gamma))}{e^{\exp(\gamma) - 1} \, \wt{\Gamma}(n+1, \exp(\gamma) - 1)} \\
    &= c_0(n,k) \frac{\wt{\Gamma}(n - k + 1, \exp(\gamma))}{\wt{\Gamma}(n+1, \exp(\gamma) - 1)}. 
\end{align*}
At this point, we consider the upper bound on the probability of interest as stated in the proposition. We have 
\begin{align*}
    \frac{\wt{\Gamma}(n - k + 1, \exp(\gamma))}{\wt{\Gamma}(n+1, \exp(\gamma) - 1)} = \frac{\int^{\infty}_{\exp(\gamma)}t^{n-k}e^{-t}dt}{\int^{\infty}_{\exp(\gamma) - 1}t^{n}e^{-t}dt} \frac{\Gamma(n+1)}{\Gamma(n-k+1)} &\leq \frac{\int^{\infty}_{\exp(\gamma)}t^{n-k}e^{-t}dt}{\int^{\infty}_{\exp(\gamma) }t^{n}e^{-t}dt} \frac{n!}{(n-k)!} \\
    & \leq \exp(-\gamma k) n^k = \exp(-\delta k \log n). 
\end{align*}
provided $\gamma \geq (1+\delta) \log n$, which concludes the proof of the upper bound. 

Regarding the lower bound, observe that in view of relation \eqref{eq:poisson_incomplete_gamma}, the ratio
of normalized incomplete Gamma functions can be expressed via the ratio of CDFs of two independent Poisson 
random variables, that is 
\begin{equation*}
\frac{\wt{\Gamma}(n - k + 1, \exp(\gamma))}{\wt{\Gamma}(n+1, \exp(\gamma) - 1)} = \frac{\p(X_{1} \leq n - k)}{\p(X_{2} \leq n)},
\end{equation*}
where $X_1$ and $X_2$ are two independent Poisson random variables with parameters
$\exp(\gamma)$ and $\exp(\gamma) - 1$, respectively. Setting
$\gamma = \log(n - k)$ yields that the right hand side is a function of the form $c_1(n,k)$ that is lower and upper bounded by
$1/4$ and $1$, respectively, as $n \rightarrow \infty$.  Taking $c(k,n) = c_0(k,n) \cdot c_1(k,n)$ yields the assertion.

\section{Proof of Proposition \ref{prop:banded}}
Similar to Eq.~\eqref{eq:basic_inequality_linear} in the proof of Theorem \ref{theo:hamming}, we have the basic 
inequality 
\begin{equation*}
-\nscp{\M{Y}}{\wh{\Pi} \bm{\mu}} + \sigma_*^2 \gamma \sum_{(i,j): |i-j| > r} \wh{\Pi}_{ij} \leq -\nscp{\M{Y}}{\Pi^* \bm{\mu}}.
\end{equation*}
In the sequel, we will show that under the stated conditions, the left hand side must exceed the right hand side unless
$\wh{\Pi}_{ij} = 0$ for all $(i,j)$ such that $|i-j| > r$. Expanding $\M{Y} = \Pi^* \bm{\mu} + \sigma_* \bm{\epsilon}$ and
re-arrangings term yields the inequality 
\begin{equation}\label{eq:shuffled_key}
\sigma_*^2 \gamma \sum_{(i,j): |i-j| > r} \wh{\Pi}_{ij} \leq \sigma_* \nscp{\wh{\Pi}\bm{\mu} - \Pi^* \bm{\mu}}{\bm{\epsilon}} = \sigma_* \sum_{i: |\wh{\pi}(i) - i| > r} \eps_i (\mu_{\wh{\pi}(i)} - \mu_{\pi^*(i)}) + \sigma_* \sum_{i: |\wh{\pi}(i) - i| \leq r}  \eps_i (\mu_{\wh{\pi}(i)} - \mu_{\pi^*(i)}). 
\end{equation}
where we have used that $\nnorm{\Pi^* \bm{\mu}}_2^2 - \nscp{\wh{\Pi} \bm{\mu}}{\Pi^* \bm{\mu}} \geq 0$. For the second term on the right hand side, the triangle inequality yields that for all indices $i$ that are summed over, we have $|\wh{\pi}(i) - \pi^*(i)| \leq 2r$. Using the Cauchy-Schwarz inequality in combination with the Lipschitz property of the underlying function, we obtain that 
\begin{align}
\sum_{i: |\wh{\pi}(i) - i| \leq r}  \eps_i (\mu_{\wh{\pi}(i)} - \mu_{\pi^*(i)}) &\leq \left(\sum_{i: |\wh{\pi}(i) - i| \leq r} \eps_i^2 \right)^{1/2} \left(\sum_{i: |\wh{\pi}(i) - i| \leq r}   (\mu_{\wh{\pi}(i)} - \mu_{\pi^*(i)})^2 \right)^{1/2} \notag \\
&\leq \frac{2r\cdot L}{\sqrt{n}} \nnorm{\bm{\eps}}_2 \label{eq:exploit_lipschitz} 
\end{align}
By standard concentration results \cite[e.g.,][$\S$2.3]{wainwright_2019}, the event $\mc{E}_1 = \{ \nnorm{\bm{\eps}}_2 \leq \sqrt{2n} \}$ holds with probability at least $1 - \exp((\sqrt{2}-1)^2/2)$. 
We now turn to the first term on the right hand side of \eqref{eq:shuffled_key}. We have the upper bound 
\begin{equation}\label{eq:crude_infinity}
\sum_{i: |\wh{\pi}(i) - i| > r} \eps_i (\mu_{\wh{\pi}(i)} - \mu_{\pi^*(i)}) \leq L \nnorm{\bm{\epsilon}}_{\infty} \cdot \textsf{card}(\{i: |\wh{\pi}(i) - i|>r \}),
\end{equation}
where we have used that $\max_{i \neq j} |\mu_i - \mu_j| \leq L$. Standard concentration results yield that 
the event $\mc{E}_2 = \{\nnorm{\bm{\eps}}_{\infty} \leq 2 \sqrt{\log n} \}$ holds with probability at least $1 - 2/n$. 
Combining \eqref{eq:exploit_lipschitz} and \eqref{eq:crude_infinity} yields that conditional 
on $\mc{E}_1$ and $\mc{E}_2$, the right hand side of \eqref{eq:shuffled_key} is upper bounded by 
\begin{equation*}
2 \sigma_* L \left(\sqrt{\log n} \, \cdot \textsf{card}(\{i: |\wh{\pi}(i) - i|>r \}) +  \sqrt{2}r\right).
\end{equation*}
At the same time, the left hand side of \eqref{eq:shuffled_key} evaluates as $\sigma_*^2 \gamma \cdot \textsf{card}(\{i: |\wh{\pi}(i) - i|>r \})$. If the expression $\textsf{card}(\ldots)$ is zero, the claim follows trivially. Otherwise,
the condition $\gamma > \frac{2L (\sqrt{\log n} + \sqrt{2}r)}{\sigma_*}$ ensures that the left hand side exceeds the 
right hand side, which is a contradiction, and hence it must hold that $|\wh{\pi}(i) - i| \leq r$, $1 \leq i \leq n$. 

The
``in particular" part of the statement then follows immediately from the triangle inequality and the Lipschitz property. 


\section{Integrated maximum likelihood estimator and overfitting}\label{app:overfitting_integrated}
In this section, it is briefly explained that under a uniform prior $p(\pi) \propto 1$, the integrated maximum likelihood estimator based on \eqref{eq:integrated_likelihood} still exhibits a tendency to overfit, in a spirit similar to what is shown in $\S$\ref{subsec:mot_example} for the maximum likelihood 
estimator of $\pi^*$. To demonstrate this point, we consider the following setup: 
\begin{align}\label{eq:overfitting_example_model}
\M{Y} | \M{X}, \Pi, \beta, \sigma_*^2 \sim N(\Pi \M{X} \beta, \sigma_*^2), \qquad p(\Pi) \propto 1, \quad  p(\beta) \propto 1,
\end{align}
and $\sigma_*^2 > 0$ fixed. The integrated likelihood corresponding to \eqref{eq:integrated_likelihood} is then given by
\begin{align*}
 L(\beta) = p(\mc{D} | \beta) = \int p(\mc{D}|\pi,\beta) p(\pi|\beta) \,d\pi
                    = \int \frac{p(\pi, \beta | \mc{D}) p(\mc{D})}{p(\pi|\beta) p(\beta)} p(\pi|\beta) \, d\pi
                    \propto \int  p(\beta | \pi, \mc{D}) p(\pi | \mc{D}) \, d\pi.   
\end{align*}
Observe that under \eqref{eq:overfitting_example_model}
\begin{equation*}
p(\beta | \pi, \mc{D}) \sim N((\M{X}^T \M{X})^{-1} \M{X}^{\T} \Pi \M{X} \beta^*, \sigma_*^2 (\M{X}^{\T} \M{X})^{-1}) , \qquad p(\pi | \mc{D}) \propto \exp\left(-\frac{\nnorm{\texttt{P}_{\Pi \M{X}}^{\perp} \M{Y}}_2^2}{2\sigma_*^2}  \right),
\end{equation*}
where $\texttt{P}_{\Pi \M{X}}^{\perp}$ denotes the projection on the orthogonal complement of the column space of
$\Pi \M{X}$. Note that $p(\pi | \mc{D})$ is high for permutations achieving good fit (overfit) to the data, and the optimization problem
$\max_{\beta} p(\mc{D}|\beta)$ will be dominated by the modes of those distributions $p(\beta|\pi, \mc{D})$ for which
the corresponding weight $p(\pi | \mc{D})$ is high. In particular, in regimes with $\textsf{SNR} = \nnorm{\beta^*}_2^2 / \sigma_*^2$ large, the maximizer of the integrated
likelihood will not substantially differ from the estimator returned by $\max_{\pi, \beta} p(\mc{D}|\beta, \pi)$ (the 
MLE in $\S$\ref{subsec:mot_example}), which is known to overfit dramatically.


\section{Metropolis-Hastings scheme for local permutations}\label{app:MH_local}
\begin{minipage}{\textwidth}
\begin{algorithm}[H]
\caption{Monte Carlo EM Algorithm for local permutations}\label{alg:MH:local_shuffle}
{\small
{\bfseries Input}: $\mc{D},  \theta, \wh{\pi}_{\text{init}}, \gamma, m, r$\par 
\textbf{Initialize} $\pi^{(0)} \leftarrow \wh{\pi}_{\text{init}}$. \par
{\bfseries for} $k=0,\ldots,m$
\begin{itemize}[leftmargin = .85ex]
\item[] Sample $i \in [n]$ uniformly at random. 
\item[] Sample $j$ uniformly from $\{\max\{i-r,1\},...,\min\{i+r,n\}\}$.
\item[] $\mathbf{If}\; |\pi^{(k)}(i) - \pi^{(k)}(j)| > r$ 
\item [] $\quad$ \texttt{invalid-mcmc-steps} $\leftarrow$ \texttt{invalid-mcmc-steps} + 1; \textbf{continue}; 
\item[] $\mathbf{end \;If}$
\item[] $\wt{\pi}(i) \leftarrow \pi^{(k)}(j)$, $\wt{\pi}(j) = \pi^{(k)}(i)$. 
\item[] $r(\wt{\pi}, \pi^{(k)}) \leftarrow \min \left\{ \frac{p(\wt{\pi}|\mc{D}, \theta;\gamma)}{p(\pi^{(k)}|\mc{D}, \theta;\gamma)}, 1  \right \}$.
\item[] Draw $u \sim U([0,1])$.
\item[] {\bf if} $r(\wt{\pi}, \pi^{(k)}) > u$: $\pi^{(k + 1)} \leftarrow \wt{\pi}$.
\item[] {\bf else}: $\pi^{(k + 1)} \leftarrow \pi^{(k)}$.
\item[] $k \leftarrow k+1$. 
\end{itemize}
{\bfseries end for} \\
{\bfseries return} $\wh{\E}[\pi | \mc{D}, \theta]$ as in \eqref{eq:MC-EM} with $m$ replaced by
$m$ - \texttt{invalid-mcmc-steps}.}
\end{algorithm}
\end{minipage}

\end{document}